%% file: main.tex
\documentclass{egpubl}
\usepackage{eg2022}
 
\ConferencePaper        
\usepackage[T1]{fontenc}
\usepackage{dfadobe}  
\usepackage[utf8]{inputenc}
\usepackage{changepage}

\usepackage{booktabs} 

\nonstopmode

\usepackage[ruled]{algorithm2e} 

\usepackage{graphicx,multirow}
\usepackage[dvipsnames]{xcolor}
\usepackage{bbm}
\usepackage{multirow}
\usepackage{booktabs}
\usepackage{amsmath,amssymb}

\ifdefined\AFfast
  \usepackage[allfiguresdraft]{draftfigure}
\fi

\SetAlFnt{\small}
\SetAlCapFnt{\small}
\SetAlCapNameFnt{\small}
\SetAlCapHSkip{0pt}
\IncMargin{-\parindent}

\usepackage{cite}  

\BibtexOrBiblatex
\electronicVersion
\PrintedOrElectronic

\usepackage{egweblnk} 

\title{Learning from Shader Program Traces}

\author[Y. Yang \& C. Barnes \& A. Finkelstein]
{\parbox{\textwidth}{\centering Y. Yang$^{1}$
         C. Barnes$^{2}$ 
         and A. Finkelstein$^{1}$
        }
        \\
{\parbox{\textwidth}{\centering $^1$Princeton University, USA \\
$^2$Adobe Research, USA
       }
}
}

\newcommand{\ignore}[1]{}
\input{0-macros}

\begin{document}

\input{figtex/result_fig_def}
\input{figtex/teaser}

\maketitle


\input{0-abstract.tex}

\section{Introduction}
\label{sec:intro}

Deep learning applications in graphics and vision typically work on images encoded as pixels in RGB color space. For images with 3D scenes, researchers have also explored augmenting the RGB data with hand-picked features like depth or surface normals~\cite{Chaitanya,vogels2018denoising}. These auxiliary features are picked based on domain expertise, and vary for different applications or programs.

This paper proposes augmenting the data from which a neural network learns with the \emph{program trace}. In software engineering, a trace refers to the record of all states that a program visits during its execution~\cite{feiler1993software,larus1993efficient}, including all instructions and data. 
We explore this idea in the context of procedural shader programs, like the one shown in \fig{fig:teaser}. 
The sequence of instructions tend to be similar from pixel to pixel, so
we rely on just the intermediate values for learning, referring to these 
as the ``program trace.''

Shader programs can be used to flexibly construct complex and even fantastical appearances by combining sequences of mathematical operations to create texture patterns, produce lighting, perturb surface normals to produce effects such as bump mapping, apply noise functions, or determine ray intersections with procedurally generated geometry~\cite{akenine2008real}.
A range of example shaders may be seen throughout this paper; many more examples are available from websites such as
\shadertoy.
Note that while the example shaders appearing here are simpler than those typical of production or games, they embody the key features that appear in production-level shaders.

Since the fragment shader program operates independently per pixel, we can consider the full program trace as a vector of values computed at each pixel -- a generalization from simple RGB. Since there are many pixels (program traces) per image, and potentially many computed images, this provides a rich source of data from which to learn.
Graphics and vision researchers have explored learning algorithms for input-output image pairs with a few auxiliary feature buffers, such as those of
Vogels~\etal~\shortcite{vogels2018denoising} on removing sampling noise,
and Xie~\etal~\shortcite{xie2018tempoGAN} on fluid super-resolution.
Such features are \emph{manually} identified by an expert on a per-shader basis.
Moreover, the extent to which these auxiliary features helps learning depends on the choice of features, the particular shader, and the learning goal. We believe other shader-specific information useful to the learner remains hidden within the program execution, and that a learning process could \emph{automatically} identify and leverage that information. Thus, we propose a learning-based approach that utilizes all of the information produced during the execution of a shader program. 
The learner could
automatically identify which features are useful, obviating the need for manual feature selection in the midst of an experimental process.

\changedh{Intuitively, learning tasks that extrapolate from a partial computation to predict the result of a full computation may benefit from learning from program traces. To illustrate its applicability,}
we introduce four applications.
Three of them work from pixel data: learning to predict low-noise output, learning to reconstruct full computation from a program with partial computation, and learning the output of a postprocessing filter.
The fourth application shows that the idea of learning from program traces can be applied to non-imagery data: it learns to simulate the position and velocity of a flock of ``boids'' \cite{reynolds:1987:CG},
which emulate flocking behavior similar to that of birds or fish.
\changedh{The performance of these applications is summarized in \fig{fig:summary_bar}.}
In most of our experiments, we train a separate model for each shader on each application. Scene specific learning is commonly used in recent work on novel view synthesis~\cite{sitzmann2019deepvoxels,thies2019neural,thies2020imageguided,mildenhall2020nerf}.
\sect{sec:multiple_shaders} describes how a single network can be trained over \emph{multiple} shaders.

The primary contribution of this paper is the idea that a shader program trace can be used as a feature vector for machine learning. Nevertheless, it is neither obvious how to use such a feature, nor that it would help in any particular application. Thus, a secondary contribution is to introduce a framework for learning from program traces, and demonstrate that it outperforms baseline methods in several applications. 
The third contribution is to investigate the relative importance of individual trace features, and how the input trace size across various trace subsampling strategies can affect the performance of the model. 

\input{figtex/summary_bar.tex}

\section{Related Work}
\label{sec:related}

\ourParagraph{Program traces in machine learning.} Program traces have  proven helpful in malware detection \cite{DBLP:journals/corr/abs-1801-02318}, program induction \cite{Reed2016NeuralP} and program synthesis \cite{chen2018executionguided,9218613}. 
Researchers have also explored using partial execution or partial rendering to synthesize graphics programs \cite{DBLP:journals/corr/abs-1804-01118} or infer parameters for procedural models \cite{NIPS2016_40008b9a}.
Instead of developing specialized learning models for a particular application, we explore a generic architecture that can learn over
a range of applications. Nevertheless, this work focuses on learning from program traces for shaders, which enjoy certain unique properties such as an emphasis on pixel outputs and an enormous degree of parallelism. 

\ourParagraph{Features for deep learning on imagery data.} Researchers have explored a variety of features beyond simple RGB as inputs to learned functions. Nalbach~\etal~\shortcite{Nalbach2017b} have made a comprehensive exploration of such features as part of a deferred shading pipeline. Xie~\etal~\shortcite{xie2018tempoGAN} consider auxiliary features including flow velocity and vorticity when learning density super-resolution for fluid simulation.
Positional encoding~\cite{mildenhall2020nerf} augments the input by applying high frequency functions to coordinate features, but does not benefit our tasks (\appx{sec:positional}).
To our knowledge, our paper is the first to propose augmenting such features with the full program trace. The benefits are that the trace of a program that computes such manually picked features inherently includes them, as well as other potentially useful information; moreover the extent to which various features are useful for a particular application and shader are discovered automatically by the learning process.

\ourParagraph{Feature space reduction.} In deep networks, an overly large feature space can exhaust memory, increase training time, or even make learning tasks harder. Researchers have explored methods to reduce the feature space by pruning whole convolutional filters \cite{DBLP:journals/corr/LiKDSG16,Luo_2017_ICCV,DBLP:journals/corr/MolchanovTKAK16}. In our method, we focus mostly on reducing the input feature space because the dimension of the program trace can be large. 
We use a method similar to that of Molchanov~\etal~\shortcite{DBLP:journals/corr/abs-1906-10771} to evaluate the importance of each trace input and 
show a trade-off between the runtime and visual fidelity as we change our feature reduction strategies. Our experimental results suggest that 
without prior execution or learning, we could find no subsampling strategy that consistently outperforms a simple uniform subsampling. However, if the shader is allowed the overhead of learning a model over the full program trace, we can select important trace features from the learned model.

\ourParagraph{Remove sampling noise.} One of our applications addresses Monte Carlo noise reduction in low-budget rendering. 
One strategy for low-noise rendering involved carefully distributing the samples, see Zwicker \etal~\shortcite{zwicker2015recent} for a survey. 
Another method uses symbolic compilation techniques to
analytically approximate the integral that produces the smoothed shader~\cite{dorn2015towards,DBLP:journals/cgf/YangB18}, but is hard to scale to complicated shaders such as the one shown in \fig{fig:teaser}.
On the other hand, learning-based algorithms train regressors such as neural networks to predict the rendering. The input to networks are usually augmented with auxiliary features~\cite{Chaitanya,vogels2018denoising,gharbi2019sample}. Unlike previous work, our approach gathers customized information per shader, and is orthogonal to learning-based denoising network design in the sense that it can be combined with an existing network.

\ourParagraph{Shader simplification.} 
As the complexity of the shader program grows, it is common to apply lossy optimization to obtain programs that only approximate the original program but with better runtime performance \cite{he2015system,genprog_simpl,wang2014automatic}.
We show experiments that explore how the trace from the simplified programs can provide information that helps to recover the missing details in the target shader. 
Thies~\etal~\shortcite{thies2019neural} identify a similar task where they learn novel view synthesis from a coarse proxy geometry. 
(The \venice\ example shown in \fig{fig:teaser} is particularly reminiscent of that work.)
Nevertheless, to our knowledge this is the first paper to propose the application of learning from a simplified shader program to restore details in the original program; and we show that using the program trace can help in this application.

\ourParagraph{Neural networks for image processing.} Researchers have investigated a variety of learning-based methods for image processing tasks, such as image enhancement and filtering~\cite{gharbi2017deep,li2018differentiable,li2017joint,wu2018fast}. 
Our postprocessing application demonstrates that the proposed method is also helpful when learning these imagery operations as postprocessing filters.

\ourParagraph{Learning simulation programs.} High quality simulation usually executes the program over many tiny time steps, which is expensive. Researchers have developed reinforcement learning based methods \cite{DBLP:journals/corr/abs-1905-04077,Kim2020_GameGan} to replace the program entirely, or execute the program at a lower spatial resolution and learn a super-resolution model \cite{DBLP:journals/corr/abs-1906-01689}. Our approach instead learns from the program's execution trace on a larger time step, and corrects the output as if the program is executed for multiple smaller steps.

\vspace{-0.5\baselineskip}

\section{Compiler and Preprocessing}
\label{sec:method}

This section introduces a compiler that can collect traces from shader programs. It translates shader programs from a domain specific language to TensorFlow code that logs the trace (\sect{sec:trace}). To stay within the hardware memory budget, the compiler also restricts the trace length to an arbitrary size cap (\sect{sec:loops}).
Collected program traces are further preprocessed before learning (\sect{sec:preprocessing}).
\sect{sec:network} describes the learning process in detail.

\vspace{-0.75\baselineskip}

\subsection{Compiler and Program Traces}
\label{sec:trace}

Our compiler takes as input an arbitrary procedural shader program written in a domain specific language (DSL) and translates it to a TensorFlow (TF) program that outputs a rendered image as well as a collected program trace. 
We embed the DSL in Python, which allows us to use Pythonic features such as operator overloading. We also include common shader operations such as trigonometric functions, dot and cross products.
For simplicity, we assume the shader program manipulates numerical scalars or vectors of known size. 
We handle branching by computing both branches of conditionals. Likewise, loops are unrolled to the maximum possible number of iterations: this limit is set by the programmer for each loop.
These are not fundamental limitations of the approach, as we experimented emulating branching and variable-length loops by writing dummy values of zero to traces in the branch/iteration not executed, and this gives visually and quantitatively identical results to our current approaches (\appx{sec:emulation}).
These policies permit us to express the trace of any shader as a fixed-length vector of the computed scalar values, regardless of the pixel location.

\vspace{-0.75\baselineskip}

\subsection{Feature Vector Reduction}
\label{sec:loops}

Large program traces can produce unnecessarily large feature vectors from which learning becomes unwieldy, or worse, exhausts memory. Loop unrolling is a common contributor to large traces, 
because the program trace would be scaled by the number of iterations.
The remainder of this section describes several strategies for reducing the size of the feature vector. All these strategies 
can be reused when targeting a different language, 
e.g. GLSL or CUDA.

\ourParagraph{Compiler optimizations.} 
Since such features would be redundant in the learning network, the compiler omits constant values, duplicate nodes in the compute graph, and neighboring nodes that differ only by a constant addition or multiplication.
The compiler also identifies common built-in functions and iterative improvement loops to eliminate trace features that are highly correlated.

Built-in functions (e.g., {\texttt sin}) should typically be treated as a black box. Our DSL provides 
widely used shader operations such as noise functions and a normal computation functor. The compiler logs only the return values of such built-in functions, not the intermediate values found when computing them. 
This is a natural choice since in principle one could trace down to a very low level such as including details about the microarchitecture, but we believe that learning will gain the most benefit if it occurs at a similar abstraction level as used by the programmer.

An iterative improvement loop repeatedly improves an approximate result to obtain a more accurate result~\cite{sidiroglou2011managing}. A commonly used iterative improvement pattern in shader prototyping is a ray marching loop that computes the distance from the camera to objects in the scene. 
Because each iteration computes a more accurate approximation than the previous iterations, the final iteration is the most informative. Therefore, the compiler will only log the trace from the final iteration of such loops. We automatically handle common cases of iterative improvement loops found in shaders by classifying loops based on a pattern matching: the output of the loop is either iterative additive or can be written as a parametric form of the iterative additive variable. 
Detailed classification rules appear in \appx{sec:classify_loop}. %
We also investigate several other strategies inspired by previous work on 
loop perforation~\cite{sidiroglou2011managing} and 
image perforation~\cite{lou2016image}. In our case, however, we always run the full computation, but simply select a subset of those computations as input to the learning task, as follows.

\ourParagraph{Uniform feature subsampling.} 
The most straightforward strategy is to subsample the vector by some factor~$n$, retaining only every  $n^{\textrm{th}}$ trace feature as ordered in a depth
first traversal of the compute graph. 
This approach tends to work well in our experiments, 
and we speculate that it does so because nearby nodes in the
compute graph tend to be related by simple computations and thus are redundant.

\ourParagraph{Other sampling schemes.} 
We explored a variety of other schemes to reduce the feature vector length,
%
\changedh{including ``clustering" based on statistical correlation, “loop subsampling” that logs features from every kth loop execution; “first or last” which only collects features from either the first or last iteration of a loop; and “mean and variance” summarize the statistics of a variable over all loop iterations.}
Yet none outperformed the above
straightforward scheme consistently enough to justify their use in our subsequent experiments. 

These options are combined as follows.
We first apply compiler optimizations, then subsample the  features with a subsampling rate that makes the trace length be most similar to a fixed target length. For all experiments, we target a length of 200, except where specifically noted such as in the simulation example. 
%
After compiling and executing the shader, we have for every pixel: a vector of dimension \tracelen: the number of recorded intermediate values in the trace.

\vspace{-0.75\baselineskip}

\subsection{Whitening the Collected Trace}
\label{sec:preprocessing}

We preprocess the traces to rescale the data to a fixed range.
Intermediate values in computed shader programs can vary over a large range:
resulting in values such as $10^{30}$, $\pm\infty$, or \emph{not a number} (NaN), even when most values of this shader computation remain near zero.
This can happen, for instance, near object silhouettes where textures have high frequency in image space.
The extreme values could cause a standard whitening technique to  fail entirely, due to say undefined mean or standard deviation where values such as $\pm\infty$ or NaN are present. Even if only finite trace values are observed at training time, standard whitening may focus too much on extreme values such as $10^{30}$, resulting in meaningful data (e.g. between $[-1, 1]$) being mapped to a very small range, and at test time, extreme values such as $\pm\infty$ or NaN can still produce  non-finite floating point values that are problematic for inference.

We thus develop a whitening method for shader program traces. We first clamp the raw program traces by collecting the statistics for the intermediate values' distribution at training time and decide the clamping threshold based on the lowest and highest $p$th percentile ($p = 5$ in practice). The clamped values are then rescaled to a fixed range [-1, 1]. More details are described in \appx{sec:preprocessing_appx}.

We evaluated the effectiveness of scaling and clamping on the denoising task (\sect{sec:denoising}) with \mandelbrot. If trained without clamping, the model will diverge to NaN even before the first iteration finishes, while training without whitening results in 12x worse perceptual error compared to our full method. These results indicate that our data preprocessing is essential in our pipeline.

\vspace{-0.75\baselineskip}

\section{Network Architecture and Training Details}
\label{sec:network}
\input{figtex/arch}

\input{figtex/result_fig_denoising}

This section briefly summarizes the training details in our experiments.
For all of our imagery applications we selected a basic architecture described in \sect{sec:architecture}.
Nevertheless, our method could be coupled with any deep learning architecture. This section thus serves as an example of how to select a network architecture and carry out training.

\vspace{-0.75\baselineskip}

\subsection{Network Architecture}
\label{sec:architecture}

Our experiments use a dilated convolutional neural network depicted in \fig{fig:architecture}, similar to that of Chen~\etal~\shortcite{Chen_2017_ICCV}.
This network architecture is used directly in our denoising (\sect{sec:denoising}) and post-processing (\sect{sec:postprocessing}) applications, and also serves as a generator model in other applications and scenarios, 
each of which relies on a GAN model: conditional spatial GAN \cite{isola2017image} for learning from a simplified shader (\sect{sec:simplified}) and temporal GAN \cite{wang2018vid2vid} for learning temporally coherent sequences (\appx{sec:temporal}). We prefer to keep a single network architecture for consistency and ease of experimentation across all applications except boids, in order to demonstrate our core idea of learning from shader program traces is beneficial across many applications, although more specialized architectures could be beneficial for certain applications like denoising (e.g.~\cite{gharbi2019sample}). Details about the GAN models are discussed in \appx{sec:gan}. The boids simulation relies on a fully-connected architecture described in \appx{sec:boids_appx}.


\subsection{Loss Functions}
\label{sec:loss}

We use a combination of pixel-wise color loss $L_c$ and perceptual similarity loss $L_p$ to encourage network output to be similar to the ground truth during training: $L_{b} = L_{c} + \alpha L_{p}$.
%
%
The parameter $\alpha$ is a weight that balances between the color and perceptual loss terms. We fix $\alpha = 0.04$ for all of our experiments. This value was chosen to roughly balance the magnitude of the gradients due to $L_{c}$ and $L_{p}$ during back-propagation.
The color term $L_c$ is simply the standard $L_2$ loss on the RGB image.
The other loss term $L_p$ uses the learned image perceptual dissimilarity metric of Zhang~et~al.~\shortcite{zhang2018perceptual}. 
This section describes the basic loss $L_{b}$ used in training. Additional details about the GAN losses can be found in \appx{sec:gan}, and the loss used in the boids simulation is described in \appx{sec:boids_appx}.

\subsection{On the Fly Training}
\label{sec:onthefly}

\vspace{0.5\baselineskip}

In training, we generate input program traces on the fly each time one is needed, rather than loading precomputed traces from disk. There are two benefits to this approach. First, precomputed traces are large, and it is typically faster to re-compute the trace, as opposed to loading from disk. Second, each time a trace is generated, we use a new randomly sampled sub-pixel location for evaluating the trace for any given pixel (a common strategy to reduce aliasing). Therefore, the input traces will generally have different values in each epoch even though we use the same ground truth solution. This approach helps the network avoid overfitting.

\newpage

\section{Evaluation}
\label{sec:eval}

This section describes experiments evaluating our method for various applications and scenarios: 
denoising pixel shaders (\sect{sec:denoising}), 
learning to reconstruct simplified shaders (\sect{sec:simplified}),
learning postprocessing effects (\sect{sec:postprocessing}), and learning non-imagery simulation programs (\sect{sec:simulation}).
We also discuss learning temporal coherence in \appx{sec:temporal}.
The architecture and training scheme in these applications includes fully connected networks, traditional CNNs and GANs, demonstrating our method's wide applicability to various deep learning models.
We report LPIPS, SSIM and PSNR for all applications in \tbl{tab:err_main},
with a performance summary shown in \fig{fig:summary_bar}:
in all cases our method outperforms the strongest baseline.

For the image processing applications,
we choose a variety of challenging combinations of shaders and geometries. 
The \bricks\ shader relies on simplex noise~\cite{perlin2001noise}.
The \trippy, \mandelbrot\ and \mandelbulb\ shaders rely on iterative fractals. 
The shaders \mandelbulb, \primitives, \oceanic\ and 
\venice\ construct complex 3D procedural geometry rendered by ray marching over a signed distance field~\cite{raymarch}.
\changedi{The shaders \bricks\ and \venice\ extract contents from texture maps.}
We adapted shaders \oceanic, \trippy, \mandelbulb\ and \venice\ from shaders with the same names at the web site 
\shadertoy, by the authors \emph{Frankenburgh}, \emph{Cras}, \emph{EvilRyu}, and \emph{reinder}, respectively,
while \primitives\ is adapted from the shader ``primitives'' by author \emph{Iq};
and the boids and fluid simulations
were adapted from
``Simple Boids'' by \emph{Saduras} and ``Chimera's Breath'' by \emph{nimitz}.

Our implementation is trained on a single GPU. For consistent timing in evaluation, we use a 4 core Intel Xeon E5-2620 v4 2.10 GHz CPU with a single Nvidia GForce RTX 2080 Ti GPU across all models. During training, we always train 400 epochs for models without a GAN and 800 epochs for models with a GAN.
Timing results reported throughout appear as speedup relative to ground truth. 
\changedh{The actual shader runtime ranges from 30ms to 21s with a median of 1.4s for full computation i.e. non-simplified shaders (\sect{sec:denoising} and \ref{sec:postprocessing}), and from 20ms to 6s with a median of 80ms for partial computation (\sect{sec:simplified}). Inference time ranges from 70ms to 0.2s with a median of 90ms.}
%
These shaders are relatively slow because they are implemented as computational graphs in TensorFlow.  They could be greatly accelerated through engineering a GLSL or CUDA implementation.
Note the shader's runtime is invariant to whether program traces are collected or not, therefore it is not a limitation to our proposed method.
In all cases, we select the model at the epoch with the lowest validation loss.
For imagery learning tasks (\sect{sec:denoising}, \ref{sec:simplified}, \ref{sec:postprocessing}),
the model trains on a dataset of 1200 tiles with $320\times320$ resolution, and 120 validation tiles in same resolution. Testing includes 30 full size images with resolution $640 \times 960$. Please refer to \appx{sec:tiled} for further details regarding our training.
All experiments presented in this section are trained per shader. We also demonstrate in \sect{sec:multiple_shaders} that multiple shaders can be trained together with a shared network and a lightweight shader-specific encoder.

\input{figtex/error_table_main}

Our strongest baseline is RGBx. It uses the same network and training as ours, but with the input features consisting of RGB color plus manually picked auxiliary features that are commonly used for learning with shader programs. 
We use normal, depth, diffuse and specular color whenever these terms are explicitly represented in the program. These corresponds to auxiliary features used in recent denoising papers ~\cite{Chaitanya,vogels2018denoising}.
%
Because the RGBx baseline generally has fewer input channels compared to our method, 
we increase the number of channels in the first convolutional layer of the baseline model such that the number of trainable weights matches that of our model.
Unlike our \emph{automatic} method, RGBx requires additional \emph{manual} expertise to pick auxiliary features for every shader program. An automatic baseline that resembles ours would be RGB, which uses only RGB color without any auxiliary features. However, RGBx always outperforms RGB, so we only compare with RGBx.

\input{figtex/result_fig_simplified}

\subsection{Denoising Fragment Shaders}
\label{sec:denoising}

Here we describe the application of removing sampling noise. Our goal is to approximate a low noise reference image collected using 1000 samples per pixel (\nspp{}). Our method is evaluated using \nspp{1}, drawn from a Gaussian spatial distribution with a standard deviation of $0.3$.

We evaluate our method and compare it against two baselines.
The first baseline is RGBx described before.
Our second baseline is supersampling. Supersampling draws a number of samples at each pixel, evaluates the shader to obtain RGB colors for each sample, and takes the mean of the colors. We supersample by choosing a constant sample budget per pixel to achieve approximately the same run time as ours, including the overhead for neural network inference.

Training for 400 epochs typically takes between 6 and 32 hours. However, the \oceanic\ shader is slower, and takes about 7 days to train.
Note that all shaders are trained using the same process over an identical architecture with a similar number of input channels; therefore the great variation in training time derives primarily from the cost of sampling from shader programs, not from learning.

In terms of the arithmetic average over all shaders, our method has a relative perceptual 
error of 67\% compared to the RGBx baseline.
A different baseline, Supersampling, is consistently worse than RGBx, with relative perceptual error ranging from 3x to 21x compared to RGBx (Appendix~\tbl{tab:error_denoising}).
We believe the dramatic improvements in relative perceptual error of our method over the baselines corresponds with the qualitatively better reconstruction of high-frequency details that we observe in the renderings
(\fig{fig:result_denoising} c-f). 
Our supplemental video shows comparisons of several of these shaders rendered with a moving camera.

\subsection{Reconstructing Simplified Shaders}
\label{sec:simplified}

We also explore a more challenging task: learning to reconstruct the appearance of a shader from its simplified variant. Shader simplification is commonly used as a lossy optimization that improves runtime while approximating the output of the original program. However, simplified programs often lose texture or geometry detail as compared with the original.
For example, the simplified versions of \mandelbrot\ and \mandelbulb\ shown in \fig{fig:result_simplified}d look obviously different from their original counterparts in \fig{fig:result_simplified}c.
We therefore propose an application that learns to recover the denoised output of the original shader from the traces of the simplified shader program sampled at \nspp{1}. 
To our knowledge, this paper is the first to propose this learning task.

We use two different techniques to simplify the shader programs: 
genetic programming simplification \cite{genprog_simpl} (on \bricks) and loop perforation \cite{sidiroglou2011managing} (on all other shaders).
%
Because the model needs to synthesize unseen texture, we use a spatial discriminator for this application, described in \appx{sec:gan}.
Training for 800 epochs takes between 10 and
60 hours. Similar to the denoising application, the great variation in training time mostly comes from generating input samples from the shader.
Our method has on average 62\% perceptual error compared to the RGBx baseline.

\subsection{Postprocessing Filters}
\label{sec:postprocessing}

\input{figtex/result_fig_post_processing}

Our method can be useful for learning not only denoising, but also applying additional image-space postprocessing filters. We implement two postprocessing filters on the CPU: an edge aware sharpening filter~\cite{Paris:2011:LLF:1964921.1964963} 
and defocus blur~\cite{rokita1993fast}. 
The network learns simultaneously to denoise and apply the postprocessing filter on the GPU.
\fig{fig:result_fig_post_processing} shows learning a defocus blur filter on \mandelbulb, and learning a sharpening filter on simplified \trippy. 
Our approach reproduces the complex effect more faithfully, as compared to RGBx, and the average relative perceptual error for ours is 74\% of that of RGBx.

\subsection{Learning to Approximate Simulation}
\label{sec:simulation}

\input{figtex/simulation_boids.tex}

Departing from learning from procedural pixel shader programs, we also explore learning to predict the future for shader programs that perform simulations. This section describes simulation of flocking behavior, while 
\appx{sec:fluid} presents learning to approximate fluid simulations.

Our shader simulates a flock of ``boids'' \cite{reynolds:1987:CG} which emulate the flocking behavior of birds or fish. Each boid has 
a 4-vector state representing 2D position and velocity.
For a flock of $K$ boids, the simulation program takes input of a $K \times 4$ tensor that represents each individual boid's initial state, 
then updates the state based on repulsion and alignment forces.
The updated state then becomes the input to the next simulation step, and so forth. The interaction between boids forms a complex flocking behavior that is difficult to predict. 
We run the ground truth simulation using a small $\delta$ step size: $2\times10^{6}$ steps with $\delta=\frac{1}{600}$s, targeting 20$\delta$ per frame at 30fps. During training we further augment the data by randomly permuting boid indices. The learning task is to correct the simulation output from a larger time step $m \cdot \delta$ in order to approximate the boids' states as if the simulation ran $m$ times for step size $\delta$. (We train with $m \in [20,64]$.)
We compare our method with two baselines: a naive baseline that directly takes the larger step simulation without any correction, and an input/output (I/O) baseline that uses the input and the output of the larger step simulation as the input to a neural network. The learning model is a combination of 1D convolution layers with 3 fully connected layers. For details please refer to \appx{sec:boids_appx}.
For reported results, we simulate 40 individual boids 
and log every program trace from the boids program.
We choose a larger program trace length than for the pixel shaders because the simulation considers all pairwise interactions between boids, and a larger program trace budget better captures these interactions.

In \fig{fig:summary_bar} and \ref{fig:simulation_boids}, we show that ours always outperforms baselines numerically and visually, even when the step size is extrapolated outside the training range ($m \in [16,128]$).
The supplemental video shows that ours recovers individual boids' interaction behaviors more faithfully with a step size of $m=20$, while the I/O baseline mainly learns the average position and velocity for the entire flock but fails to recover a reasonable distance between the boids.


\section{Trace Analysis}
\label{sec:analysis_trace}

This section presents a series of analyses that help to understand how program traces are beneficial for learning. We start by analyzing which trace features are contributing the most to a learned model. 
Based on trace importance, we then investigate which subset of the trace can be used for learning. We empirically find that if one cannot afford to first execute and learn from the full shader trace, then the Uniform subsampling used throughout \sect{sec:eval} always gives reasonable performance, and we were not able to find any strategy that consistently outperforms Uniform. However, if one is able to train an additional inital network that first uses the full program trace, then we can do better than Uniform, using a strategy that we call Oracle that selects  important features.
Finally, we show that multiple shaders can be trained together with a shared denoising network and a lightweight shader-specific encoder.

\subsection{Which Trace Features Matter in a Learned Model?}
\label{sec:trace_importance}

We characterize the importance of the trace features by quantifying the change in training loss when removing each of the trace inputs. Inspired by Molchanov~\etal~\shortcite{DBLP:journals/corr/abs-1906-10771,DBLP:journals/corr/MolchanovTKAK16}, we used the first order Taylor expansion to approximate importance of each input trace feature.
Specifically, for a model trained with loss $L$ and trace length \trace, the importance score $\Theta$ of the input trace feature $\mathbf{z}^l$ ($l = 1, ..., \trace$) with image dimension \rectdim{$M$}{$N$} across $K$ examples is:
\begin{equation}
    \Theta (\mathbf{z}^l) = \frac{1}{K} \sum_{k=1}^K |\frac{1}{M\cdot N} \sum_{m=1}^M \sum_{n=1}^N \frac{\partial L}{\partial \mathbf{z}_{m, n}^l} \cdot \mathbf{z}_{m, n}^l|
    \label{eqn:importance_score}
\end{equation}

We evaluate \Eqn{eqn:importance_score} on the denoising model for two shaders: \bricks\ and \mandelbrot. Only a small fraction of the trace results in a very high importance score.
We manually inspect what the top 10\% most important trace features represent and verified that the learned importance corresponds to human intuition. 
For example in \bricks, we found the most important traces include features that determines the distance to the nearest brick edges and the Boolean condition that decides whether the pixel is insde the mortar: this helps prevent edges from being broken. In \mandelbrot, we found the trace that controls the complex number computation for almost every iteration are among the most important features: a breakdown of such information at each level could help the model to better denoise between nearby structures.

\subsection{Which Subset of the Trace to Use for Learning?}
\label{sec:which_subset}

As discussed in \sect{sec:loops}, program traces can be arbitrarily long, and we could input only a subset of the trace for efficient learning and inference, such as Uniform subsampling used in \sect{sec:eval}.
Therefore, a natural question to ask is: given a fixed input trace length budget, what subsets of the program trace are good for learning?
The best way to answer this question is to enumerate all possible subsets of the program trace and train a separate model for each. 
However, for a shader program that has \trace\ traces before subsampling and a fixed input budget \tracelen, this strategy will introduce combinatoric $\binom{\trace}{\tracelen}$ learning tasks,  which is intractable.

To investigate how different subsets of the trace could affect learning in a practical fashion, we propose subsampling strategies we call Oracle and Opponent. Both the Oracle and Opponent strategies are based on the feature importance score (\sect{sec:trace_importance}) from a Full Trace model trained with all of the program trace. Oracle always chooses the traces that have the highest importance scores, while Opponent always chooses the ones with the lowest scores. In an analogy to the lottery ticket hypothesis~\cite{frankle2018lottery}, we hypothesize that the Oracle  exploits a winning ``lottery ticket" found within the Full Trace model, and selects out the relevant trace subset: a ``lottery ticket trace." The Opponent likewise selects losing tickets.

\input{figtex/trace_subsample_comparison}

To better understand the trade-offs associated with the subsampled trace length, we experimented with varying trace lengths using Opponent, Uniform, and Oracle subsampling and compare them with the RGBx baseline, as shown in \fig{fig:trace_subsample_comparison}. 
For each shader, the trace is subsampled by a relative sample budget compared to the full program trace length \trace\ (\eg\ \tracelen = \trace/2, \trace/4). Under a fixed budget \tracelen, in most cases the inference error decreases in the ordering of Opponent, Uniform, Oracle. This corresponds to our intuition because Oracle selects traces that are beneficial to training based on prior knowledge from the Full Trace model, and similarly Opponent selects traces that are unimportant based on the same prior knowledge.
Statistically, our hypotheses that Uniform outperforms Oracle, and Opponent outperforms Uniform each have p-values $7.2 \times 10^{-4}$ and $5.9 \times 10^{-3}$, respectively. These are smaller than a threshold of 0.025 determined by correcting the traditional p threshold of 0.05 for the two comparisons, so we conclude that the ordering \textit{Oracle outperforms Uniform outperforms Opponent} is significant. For details please see \appx{sec:stat_subsample}.

It is also worth noting that even when \tracelen\ is small (\eg\ the leftmost two data points in the plots corresponds to \tracelen\ below 50), the extra information from the program trace can still substantially reduce the relative perceptual error without significant extra cost in inference time. Because the x-axis in the plot is on a log scale, the actual performance gain would have a more drastic slope starting from RGBx to a small \tracelen.
Additionally, the current comparison is advantageous to RGBx 
as its learning capacity matches that of the Full Trace model as discussed in \sect{sec:eval}, which is more capacity than any of the subsampled models in \fig{fig:trace_subsample_comparison}.

In practice, subsampling strategies can be chosen based on resources allowed for training and inference. If there is no limit at all, training a model with the Full Trace can always give the best performance. If \tracelen\ is only limited by inference time, but extra cost and memory can be permitted during training, one could use the Oracle strategy. However, when training also becomes a practical concern,
our results suggest that without actually learning from the full trace in advance, 
there may not be a single subsampling strategy that could consistently outperform all others,
as discussed in \sect{sec:loops}.
Thus, Uniform subsampling provides an effective proxy that follows the performance of Oracle, and always outperforms the worst-case scenario Opponent. 

\subsection{Can Multiple Shaders be Learnt Together?}
\label{sec:multiple_shaders}

In this section, we explore whether part of the model can be shared across shaders with the same task. 
Because program traces are unique per shader, we propose to train a separate shallow encoder for each of the shaders, followed by a task-specific model shared across shaders. The setup is similar to the source-aware encoder by Vogels \etal ~\shortcite{vogels2018denoising}.

Four shaders (\mandelbrot, \mandelbulb, \primitives, and \trippy) are trained together for the denoising task. The encoder consists of four 1x1 convolutional layers, where the first layer outputs \reducedim\ channels and the rest output 48 channels. In our method, $\reducedim = 48$ while in the RGBx baseline \reducedim\ varies similarly as in \sect{sec:eval}. The encoder is identical to the four 1x1 convolutions that analyze the input program trace in \fig{fig:architecture}. The 48-dimensional encoding then inputs to a shared denoising network, whose architecture is identical to \fig{fig:architecture} excluding the four initial 1x1 convolutions. 
All four shaders use Uniform subsampling to bring their \tracelen\ to be closest to 200. Training alternates between the 4 shaders after each epoch, and each shader is trained for 400 epochs. 

We report the error statistics for the shared model in \tbl{tab:err_main}. Ours has on average 60\% perceptual error compared to the RGBx baseline. Although one might expect this experiment to benefit the RGBx baseline as the RGBx features are more similar, in fact, ours outperforms RGBx in all cases.

\section{Conclusion, Limitations \& Future Work}

This paper proposes the idea of learning from shader program traces. It evaluates this idea in a range of learning contexts:  denoising, simplified shaders, postprocessing filters and simulation. We describe a compiler that can produce program traces suitable for learning, as well as practical considerations like how to handle large traces and how to process the trace data to make it amenable to learning. Our method is agnostic to the learning architecture, loss function and training process; however, we also discuss a particular set of these  that worked well in our experiments. We evaluate our method on a range of shaders, over which it compares favorably with baselines. We also analyze which features are important in the trace, and explain how one can select subsets of the trace for learning.

\changedh{Links to our code, data and supplemental video may be found at our project page:}
\url{https://pixl.cs.princeton.edu/pubs/Yang_2022_LFS/}

Our method has several limitations, which offer potential avenues for future work.
First, as with many neural network based approaches, the inference time is not negligible. For example, for the denoising task, simple, fast shaders can draw sufficiently many samples via supersampling so as to outperform inference. Likewise for the simplified shader tasks, one could use the time budget for network inference to instead compute multiple samples of the original more expensive shader. Future research might address this by developing specialized networks that are more efficient for inference, along similar lines as the method of Gharbi~\etal~\shortcite{gharbi2017deep}. Another limitation of our TensorFlow implementation is that the operation of collecting program traces and concatenating them into one single tensor is not particularly efficient, and is a major cause of the inference time increasing with the trace length (\fig{fig:trace_subsample_comparison}). Additionally, TensorFlow is efficient for deep learning models, but less efficient for shader computations. 
\changedh{For example, for shaders with complex BRDFs, branching generated by varying number of ray bounces per pixel may become a major bottleneck in TF.}
We believe further engineering efforts could alleviate these bottlenecks, for example by using compiled GLSL as a frontend renderer prior to the inference process.

The experiments described this paper were performed using computer graphics shaders. Future work could explore how well the ideas introduced herein generalize to other kinds of programs that can rely on (and tolerate) approximate solutions, for example those relying on stochastic algorithms or Markov-like decision processes.

\newpage
\bibliographystyle{eg-alpha-doi}
\bibliography{paper}

\clearpage



\appendix

\section{Classifying Iterative Improvement Loops}
\label{sec:classify_loop}

Here we present our detailed classification rules for iterative improvement loops. For a loop variable $X$ at iteration $n$, we will denote its value as $X_n$.

\newtheorem{definition}{Definition}

\begin{definition}
    A loop variable $X$ is iterative additive if it matches the following pattern or its equivalent forms:
    
    \begin{equation}
        X_n = X_{n-1} + Z
    \end{equation}
    
    Here $Z$ can be any arbitrary variable.
\end{definition}

\begin{definition}
    A variable $Y$ is dependent on an iterative additive variable $X$ if it matches the following pattern or its equivalent forms:
    
    \begin{equation}
        Y_n = \text{select}(cond, Y_{n-1}, f(X_n, X_{n-1}, C))
    \end{equation}
    
    Here, $cond$ is an arbitrary Boolean variable, $f$ is an arbitrary function, and $C$ is a variable computed outside the loop, i.e. $C$ can be viewed as constant inside the loop.
\end{definition}

\begin{definition}
    A loop variable $X$ is an output variable if for any iteration $n$, its value $X_n$ is used outside the loop.
\end{definition}

\begin{definition}
    A loop is classified as an iterative improvement loop if all of its output variables are either iterative additive or are dependent on an iterative additive variable.
\end{definition}

\section{Details for Whitening the Collected Trace}
\label{sec:preprocessing_appx}

For each intermediate value in the program trace, we clamp extreme values and rescale others by the following process.

\newcommand{\perclo}{\ensuremath{P_0}}
\newcommand{\perchi}{\ensuremath{P_1}}

We clamp extreme values by collecting the statistics for the intermediate values' distribution at training time. For each intermediate value, we first decide whether its distribution merits clamping. If we detect that the distribution has only a small number of finite, discrete values (10 or fewer), we do not apply clamping to the corresponding intermediate value. For the rest of the intermediate values, we first discard infinite values and then find from their distributions the lowest and the highest $p$th percentiles, denoted \perclo\ and \perchi, and use these to compute clamping thresholds. 
Next we clamp all values to the range 
$[\perclo - \gamma(\perchi - \perclo), ~~~ \perchi + \gamma (\perchi - \perclo)]$.
We also set NaN values to the low end of this range.
Empirically, we found in our experiments that $p = \changedi{5}$ and $\gamma = 2$ work well, and we use these values for all results.
Finally, for each intermediate feature, we rescale the clamped values to the fixed range [-1,1], and record the corresponding scale and bias used. In both training and testing, the collected program traces are used directly by applying the same precomputed scale and bias, but the values will be clamped to range [-2, 2] to allow data extrapolation.

\section{Generating the Dataset}
\label{sec:tiled}

Our experiments 
generates the dataset from 800 images for training, 80 images for validation and 30 images for testing (each \rectdim{960}{640}). Although this training set size is small relative to typical deep learning tasks, we address this concern in \sect{sec:onthefly}. 
The training images are generated with random camera poses, while testing images are divided into two groups: 20 \emph{similar} distance images with camera pose sampled from the same distribution as the training set, as well as 10 \emph{different} distance images that are closer or further than the training set. 
\changedh{For some shaders, (\trippy, \mandelbrot, \mandelbulb, \venice\ and \oceanic), a periodic time parameter also changes the shader appearance, which is sampled from the same distribution for both training and testing datasets.}

We find it beneficial to further divide the training and validation set into tiles. 
One advantage is that certain features in the shader may be visually salient to humans, so we can emphasize such features to ensure they are learned well. In principle this could be accomplished with automatic saliency models (e.g.~\cite{kummerer2017understanding,mit-saliency-benchmark,MLNet_Saliency}). However, off-the-shelf saliency models are trained for natural imagery whereas our shaders are non-photorealistic, and therefore we combine both a saliency model \cite{MLNet_Saliency} and a traditional Laplacian pyramid  representation to robustly and automatically select salient tiles. Another benefit of tiled training is that it reduces memory, and it also accelerates convergence, because we can use larger mini-batches with more varied content within the same GPU memory to obtain a gradient estimator with lower mean squared error.

We sample training and validation tiles as follows.
We first generate saliency maps for each of our 800 training images \changedi{and 80 validation images} using Cornia~et~al.~\shortcite{MLNet_Saliency}. Saliency models usually incorporate a center bias that tends to give lower saliency scores to pixels closer to image boundaries. This behavior is not ideal for our framework because our training images are generated from randomly sampled camera poses so that salient content could appear anywhere in the image. Therefore, we run the saliency model on images with an extended field of view (each \rectdim{1280}{960}) where the center patches of size \rectdim{960}{640} are our original training images. This allows every pixel in the original training dataset to be away from image boundaries to avoid center bias in the resulting saliency maps.

We then subdivide each of \changedi{the training and validation} images into six \squaredim{320} tiles. For each tile, we estimate its intensity on low, middle and high frequencies by taking the average over its first, third, and fifth level of the Laplacian pyramid~\cite{burt1983laplacian}. Together with the average saliency score, these four metrics can be combined to robustly sample salient and interesting tiles for learning.

Next, we use identical sampling rules to sample one-quarter of the sampling budget from each of the four metrics. For each metric, we rank the tiles according to their associated score and only sample from the tiles whose score is within the top 25\% nonzero scores. The score of the qualified tiles will further be normalized to [0, 1], and each tile will be sampled with a probability proportional to the normalized score.

Apart from the rules described above, we find it helpful to also include a small portion of constant color tiles in the training dataset, e.g. the black background in \bricks\ \fig{fig:result_denoising}. These uninformative and constant color tiles can be easily selected from a low color variance threshold. Although some salient tiles already contain both informative and uninformative regions, they are usually close to object silhouettes and could still pose challenges when extrapolating to uninformative regions far away from the object. 

We sample a total of 1200 tiles for training and 120 tiles for validation. If the shader does not contain constant color tiles, all of the sampling budget will be used to equally sample from the 4 saliency metrics described above. Otherwise, only 95\% of the sampling budget will be sampled from saliency, and another 5\% will be sampled from low color variance tiles. Testing still relies on 30 full images.

\section{Details for GAN Models}
\label{sec:gan}

Our spatial GAN model is a conditional GAN, where the conditional labels are the RGB channels of the \nspp{1} rendering from the shader program, denoted as $c_x$. Because $c_x$ is already part of the program trace, we directly use the model from \fig{fig:architecture} as our generator and the generator's output is naturally conditioned on $c_x$. We then train the model to match the ground truth denoted as $c_y$.
Additionally, we used a patchGAN architecture similar to that of Isola \etal~\shortcite{isola2017image} with receptive field \squaredim{34} as our discriminator $D$.

Our temporal GAN model uses a similar architecture as the spatial GAN with modifications following \cite{wang2018vid2vid}.
The generator is conditioned on imagery from three consecutive frames: the current predicted frame and the two previous ones. This involves five 3-channel images as conditional labels: shader RGB output from all three frames plus the generator's output from the two previous frames. 
Because neither the shader output nor the generator output from the previous two frames is part of the program trace for the current frame, we modified the generator architecture in \fig{fig:architecture} to concatenate the additional four conditional label images \emph{after} the feature reduction layer. The rest of the architecture remains the unchanged.
We use the same discriminator architecture as for our spatial GAN, but it takes an input of sequences of frames and their corresponding conditional labels.

\changedi{We now introduce the variation on the basic loss function that incorporates the GAN loss.} 
%
We use a modified cross entropy loss \cite{DBLP:journals/corr/Goodfellow17} for both spatial and temporal GAN models. 
Our spatial GAN model is conditioned on the RGB channels of the shader program $c_x$ to approximate the distribution of the ground truth $c_y$, while our temporal GAN loss is applied to sequences $\widetilde{c_x}$ and $\widetilde{c_y}$.
The training objective (that we minimize) for generator $L_G$ and loss for spatial discriminator $L_{D_S}$ can be expressed as:

\begin{equation}
    \begin{split}
        L_G = & \ \ \ L_{b} - \beta \ \mathbb{E}_{c_x}{\log(D_{S}(G(c_x), c_x))} \\
        L_{D_S} = & -\mathbb{E}_{c_x, c_y}{\log(D_S(c_y, c_x))} \\ & -\mathbb{E}_{c_x}{\log(1 - D_S(G(c_x), c_x))}
    \end{split}
    \label{eqn:spatial_GAN}
\end{equation}

\noindent
Similarly, the training objective on temporal sequences for generator $L_G$ and temporal discriminator $L_{D_T}$ can be expressed as:

\begin{equation}
    \begin{split}
        L_G = & \ \ \ L_{b} - \beta \ \mathbb{E}_{\widetilde{c_x}}{\log(D_{T}(G(\widetilde{c_x}), \widetilde{c_x}))} \\
        L_{D_T} = & -\mathbb{E}_{\widetilde{c_x}, \widetilde{c_y}}{\log(D_T(\widetilde{c_y}, \widetilde{c_x}))} \\ & -\mathbb{E}_{\widetilde{c_x}}{\log(1 - D_T(G(\widetilde{c_x}), \widetilde{c_x}))}
    \end{split}
    \label{eqn:temporal_GAN}
\end{equation}

\noindent
The parameter $\beta$ is a weight that balances between the GAN loss and the regular color and perceptual loss. In all our experiments with GAN loss, we fix $\beta = 0.05$ to roughly balance the magnitude of gradients from all loss terms.
Note in \eqn{eqn:temporal_GAN} we did not include spatial discriminators for simplicity. But it is possible to combine both \eqn{eqn:spatial_GAN} and \eqn{eqn:temporal_GAN}. For example, in \appx{sec:temporal}, we trained on both discriminators to produce a temporally coherent model for simplified shaders.

We also skip the back-propagation on the GAN loss for any mini-batch with constant color to avoid training instability.

\section{Details for the Boids Simulation}
\label{sec:boids_appx}

The boids simulation \sect{sec:simulation} works with non-imagery data, and we therefore uses a combination of 1D convolution and fully connected layers for that learning task. 
The input to the network has size $B \times \tracelen$ where 
$B$ represents the number of boids (40 in our experiments) and
$\tracelen$ represents either the length of the program trace in our method or 8 for the I/O baseline. 
We first reduce the dimensionality of the trace to \reducedim\ using a 1D convolution with kernel size one, followed by 3 additional 1D convolutions with kernel size one and 48 output channels. This is an analogy to the 2D feature reduction layer and 1x1 convolutions described in \fig{fig:architecture}, where $\reducedim = 48$ for our method and $\reducedim = 1173$ for the I/O baseline to match the number of trainable weights in both models.
We then flatten the $B \times 48$ tensor as an input to a 3 layer fully connected network, where each layer has 256 hidden neurons, and an output fully connected layer with the number of neurons being $B \cdot 4$, representing the output state for each boid.

We learn a four-channel state (2D position and velocity) for each boid, rather than RGB. Therefore we use only $L_2$ loss on these coordinates after separately normalizing position and velocity over the training set.

\section{Training Temporally Coherent Sequences}
\label{sec:temporal}

\input{figtex/error_table_temporal}

\input{figtex/result_fig_temporal}

\input{figtex/simulation_fluid}

\input{figtex/error_table_denoising}

Temporal coherence in a graphics or vision context refers to there being a strong correlation between each frame and the next. Training only on individual images can introduce temporal incoherence for rendered video. One straightforward fix would be to apply a temporal filter to the output sequences to blur out the noise. Alternatively, we implemented a temporal discriminator to directly train temporally coherent sequences using a training scheme similar to that of Wang~\etal~\shortcite{wang2018vid2vid}. 
Each frame in a sequence is synthesized conditioned on two previous frames. In training, frames are synthesized in groups of six consecutive frames, relying on eight-frame ground truth sequences to be able to bootstrap the initial frame. 
We train temporally coherent sequences both for the task of denoising and learning from simplified programs, and compare with an RGBx baseline as in \sects{sec:denoising}~\&~\sectnum{sec:simplified}. A summary of quantitative error is shown in \tbl{tab:err_temporal}. In all cases ours outperforms the RGBx baseline, and produces a more temporally coherent sequence than their non-temporal counterparts (\sects{sec:denoising}~\&~\sectnum{sec:simplified}) while retaining similar visual quality in still images. 
We additionally verify that the temporal models generate more temporally stable sequences by computing the perceptual loss of 2 adjacent frames. For each of the 30 test sequences, we use the last two frames of the length 30 sequence and average the score across ten renders with different random seeds. We then average the score across the test dataset and compare between our temporal and our non-temporal models. In all cases, the temporal model has a lower error between adjacent frames. The temporal models have 94\% perceptual error relative to the non-temporal models on average and 80\% in the best case.
Our supplementary video does not present temporally coherent animation as a separate application, but rather shows this training scheme in the denoising and simplification applications.
\fig{fig:result_temporal} shows an example where our method generalizes better to longer sequences than the RGBx baseline. Our result correctly learns both temporal coherence as well as the complicated structure in each individual frame, whereas the RGBx baseline introduces additional color artifacts in the output. The video shows even longer sequences (180 frames).

\section{Branching and Loop Emulation}
\label{sec:emulation}

As discussed in \sect{sec:trace}, our compiler currently handles conditional execution by simply evaluating both branches and unrolls loops to its maximum possible iteration. Variable-length loops are handled using a user-given compile-time pragma specifying a ceiling on the possible number of loop iterations: it is common to have such ceilings on iteration counts in shader programs because of the need to maintain consistent shading performance. Values from unused iterations are replaced with the values from the final computed iteration.
We made these choices because they are much easier to implement in TensorFlow. However, in a practical application, shaders would typically be compiled to code that takes either branch or
exits the loop early based on a termination condition. 
Therefore, we did an experiment to determine what would be have been the effect of handling branches and loops the traditional way. 
For branching, we simply wrote dummy values of zero to traces in the branch not taken.
We applied such branch emulation to a shader called \texturemap\ which---similar to aspects of \venice\ in \fig{fig:teaser}---uses 
a conditional statement to select a texture based on whether a ray has hit a plane.
For loops, we wrote zero values to traces after the loop termination condition is met, and applied the emulation to \mandelbrot.
In both cases we found that the emulation gives results that are visually and quantitatively identical to our compiler’s implementation.

\section{Comparison with Positional Encoding}
\label{sec:positional}

Positional encoding \cite{mildenhall2020nerf} can be viewed as a general method to augment input to learning that is agnostic to the input data's generation process. It applies high-frequency functions to positional features such as 3D coordinates.
Because many shaders involve computing intermediate values that vary spatially in ways that cannot easily be captured via positional encoding, and some of them will be important to the learner,
we believe our method offers an improvement over positional encoding for most shaders and most applications.
To evaluate this hypothesis, we tested two applications (denoising in \sect{sec:denoising} and simplification in \sect{sec:simplified}) $\times$ two shaders (\mandelbrot\ and \trippy), adding explicit positional encoding features as described by \cite{mildenhall2020nerf} to both the RGBx baseline and our method. 
On average across the four cases, we found that the addition of positional encoding features did not measurably change PSNR values. 
Moreover, the addition of positional encoding features increased the perceptual error of both RGBx and ours by 4\% on average. 
Therefore, we verified our hypothesis that in the context of our applications and shaders, learning does not benefit from positional encoding.

\section{Fluid Simulation}
\label{sec:fluid}

Although our method is beneficial in all the previously described experiments, we also find a null result for our second simulation example: a 2D fluid simulation.
The state of the simulation on a 2D grid can be viewed as a 7D feature: 3D for RGB color of the fluid and 4D for internal states: velocity, density and vorticity. The simulation takes input of the 7 channel fluid state, solves the Navier-Stokes equation with a hard-coded external force to compute the new internal state, then applies color advection on image space to output the new 7D state. The  color advection step controls the trade-off between how fast the fluid propagates and how accurate the simulation is. We ran the simulation with step size $\delta$ as ground truth. The learning task is to run the simulation at a coarser step $10\delta$, and predict the intermediate states in between the 10 steps as if they were run at the fine scale simulation with step size $\delta$.

We use the same architecture as in \sect{sec:denoising} for this task and compare our method with an I/O baseline that takes the initial and output fluid states as learning features. While our method is marginally numerically better than the baseline (ours has 92\% L2 error and 96\% perceptual error compared to the baseline), the visual quality of the two methods is almost identical. We hypothesize that this learning task is not suitable for our method because it is relatively simple and lacks complicated hidden state: the neural network can easily approximate solving the Navier-Stocks equation given initial and output states. Additionally, because the fluid states change slowly even after 10 simulation steps, the network can easily hallucinate a reasonable correction using the initial state as a good starting point, therefore, the baseline features already suffice. In \fig{fig:simulation_fluid} we show both the baseline and our method can reasonably approximate the reference with almost identical results.

\section{Statistical Evidence for Subsampling Strategies}
\label{sec:stat_subsample}

In this section, we provide statistical evidence for our findings when investigating trade-offs between different subsampling strategies and subsampling budgets described in \sect{sec:which_subset}.
Our first null hypothesis makes the following assumption on the performance between Uniform and Oracle subsampling: the ratio of relative error between Uniform and Oracle ($\mu_0$) is less than or equal to $1$. This hypothesis has a p-value of $p_0 = 7.2 \times 10^{-4}$.
Similarly, we propose another null hypothesis regarding the performance between Opponent and Uniform subsampling: the ratio of relative error between Opponent and Uniform ($\mu_1$) is smaller than or equal to $1$, which has a p-value $p_1 = 5.9 \times 10^{-3}$.
If we choose a significance level of $0.05$ and apply Bonferroni correction over the 2 hypotheses, we have both $p_0 < 0.025$ and $p_1 < 0.025$, indicating significant evidence 
that Oracle outperforms Uniform ($\mu_0 > 1$) and Uniform outperforms Opponent ($\mu_1 > 1$).
These statistics are computed using all possible \tracelen\ available for all 4 shaders: $\tracelen \in [\trace/2, \trace/4, \trace/8, \trace/16]$.


\end{document}

%% file: 0-macros.tex
\newlength{\h}

\newcommand{\ignorethis} [1] {}

\newcommand{\sectnum    } [1] {\ref{#1}}

\newcommand{\sect       } [1] {Section~\sectnum{#1}}
\newcommand{\sects      } [1] {Sections~\sectnum{#1}}
\newcommand{\appx       } [1] {Appendix~\sectnum{#1}}

\newcommand{\tbl}[1]{Table~\ref{#1}}

\newcommand{\fignum     } [1] {\ref{#1}}
\newcommand{\fig        } [1] {Figure~\fignum{#1}}

\newcommand{\eqnnum     } [1] {\mbox{(\ref{#1})}}
\newcommand{\eqn        } [1] {equation~\eqnnum{#1}}
\newcommand{\Eqn        } [1] {Equation~\eqnnum{#1}}

\newcommand{\ourParagraph}[1]{\textit{#1}}

\definecolor{verydarkgreen}{rgb}{0,0.2,0}
\definecolor{verydarkorange}{rgb}{0.4,0.2,0}
\definecolor{verydarkyellow}{rgb}{0.55,0.55,0}

\ifdefined\ShowNotes
\newcommand{\Connelly}[1]{\colornote{darkgreen}{CB:}{#1}}
\newcommand{\Adam}[1]{\colornote{dblue}{AF}{#1}}
\newcommand{\Yuting}[1]{\colornote{maroon}{YY}{#1}}
\newcommand{\todo}[1]{\colornote{darkgreen}{Todo}{#1}}

\newcommand{\changedh}[1]{{\color{verydarkgreen}{#1}\normalfont}}
\else
\newcommand{\Connelly}[1]{}
\newcommand{\Adam}[1]{}
\newcommand{\Yuting}[1]{}
\newcommand{\todo}[1]{}

\newcommand{\changedh}[1]{{#1}}
\fi

\ifdefined\ShowColor
\newcommand{\changedj}[1]{{\color{dblue}{#1}\normalfont}}
\newcommand{\changedi}[1]{{\changedj{#1}}}
\else
\newcommand{\changedj}[1]{{#1}}
\newcommand{\changedi}[1]{{#1}}
\fi

\newcommand{\tablegap}[0]{\\[-0.05cm]}

\ifdefined\ShowNotes
  \newcommand{\colornote}[3]{{\color{#1}\textbf{#2: #3}\normalfont}}
\else
  \newcommand{\colornote}[3]{}
\fi

\definecolor{darkred}{rgb}{0.7,0.1,0.1}
\definecolor{darkgreen}{rgb}{0.1,0.7,0.1}
\definecolor{verydarkgreen}{rgb}{0.0,0.5,0.0}
\definecolor{verydarkblue}{rgb}{0.0,0.0,0.7}
\definecolor{cyan}{rgb}{0.7,0.0,0.7}
\definecolor{dblue}{rgb}{0.2,0.2,0.8}
\definecolor{maroon}{rgb}{0.76,.13,.28}
\definecolor{burntorange}{rgb}{0.81,.33,0}

\definecolor{uhref_color}{HTML}{000066}

\newcommand\ulinec{\bgroup\markoverwith
      {\textcolor{uhref_color}{\rule[-0.5ex]{2pt}{0.4pt}}}\ULon}

\newcommand{\nsfurl}[1]{\ulinec{\small{\color{uhref_color}{\url{#1}}}}}

\newcommand{\beginintegraltable}[2]{
\begin{tabular}{|l|l|}
	\hline
	Function $f(x)$ &
	Bandlimited with #1 kernel: $\hat{f}^{(#2)}(x)$ \\
	\hline
}
\newcommand{\finishintegraltable}{
	\hline
	\end{tabular}\\ \\
}

\makeatletter
\def\thickhline{%
  \noalign{\ifnum0=`}\fi\hrule \@height \thickarrayrulewidth \futurelet
   \reserved@a\@xthickhline}
\def\@xthickhline{\ifx\reserved@a\thickhline
               \vskip\doublerulesep
               \vskip-\thickarrayrulewidth
             \fi
      \ifnum0=`{\fi}}
\makeatother

\newlength{\thickarrayrulewidth}
\newcommand{\etal       }     {{et~al.}}

\newcommand{\eg         }     {{e.g.}}

\newcommand{\trace   }{\ensuremath{\mathcal{T}}}
\newcommand{\tracelen}{\ensuremath{\mathcal{N}}}
\newcommand{\reducedim}{\ensuremath{\mathcal{K}}}

\newcommand{\shadername}[1]{{\small\texttt{#1}}}
\newcommand{\bricks}{\shadername{Bricks}}

\newcommand{\mandelbulb}{\shadername{Mandel-bulb}}
\newcommand{\mandelbrot}{\shadername{Mandelbrot}}
\newcommand{\oceanic}{\shadername{Oceanic}}
\newcommand{\primitives}{\shadername{Gear}}
\newcommand{\gear}{\shadername{Gear}}
\newcommand{\trippy}{\shadername{Trippy}\,\shadername{Heart}}
\newcommand{\texturemap}{\shadername{Texture}\,\shadername{Maps}}
\newcommand{\venice}{\shadername{Venice}}

\newcommand{\shadertoy}{\href{https://www.shadertoy.com/}{\small\texttt www.shadertoy.com}}

\newcommand{\rectdim}[2]{{{#1}$\times${#2}}}
\newcommand{\squaredim}[1]{{\rectdim{#1}{#1}}}

\newcommand{\spp}{{\textsc{spp}}}
\newcommand{\nspp}[1]{{{#1}\,\spp}}

%% file: figtex/result_fig_def.tex
\newcommand{\ResultsFigStart}{
\setlength{\tabcolsep}{1pt}
\setlength{\h}{1.15in}
\begin{tabular}{cccccc}
}

\newcommand{\ResultsFigWithHeader}[6]{
\ResultsFigStart
(a) Reference & (b) Our result & (c) Reference & (d) Ours & (e) RGB+Aux & (f) Supersample\\
\addlinespace[4pt]
\ResultsFigEnd{#1}{#2}{#3}{#4}{#5}{#6}
}

\newcommand{\ResultsFigNoHeader}[6]{
\addlinespace[4pt]
\ResultsFigStart
\ResultsFigEnd{#1}{#2}{#3}{#4}{#5}{#6}
}



\newcommand{\ResultsFigStartTeaser}{
\setlength{\tabcolsep}{1pt}
\setlength{\h}{1.03in}
\begin{tabular}{cc@{\hskip 4pt}cc@{\hskip 4pt}cc@{\hskip 4pt}cc}
}

\newcommand{\ResultsFigEndTeaser}[5]{
\includegraphics[height=\h]{result_figs/#2_gt_box.png} & \includegraphics[height=\h]{result_figs/#2_gt_zoom.png} & \includegraphics[height=\h]{result_figs/#2_input_box.png} & \includegraphics[height=\h]{result_figs/#2_input_zoom.png} & \includegraphics[height=\h]{result_figs/#2_RGBx_box.png} & \includegraphics[height=\h]{result_figs/#2_RGBx_zoom.png} & \includegraphics[height=\h]{result_figs/#2_ours_box.png} & \includegraphics[height=\h]{result_figs/#2_ours_zoom.png} \tablegap
\multicolumn{2}{c}{\small {#1}} &
\multicolumn{2}{c}{\small {#3}} &
\multicolumn{2}{c}{\small {#4}} &
\multicolumn{2}{c}{\small {#5}}
\end{tabular}
}

\newcommand{\ResultsFigTeaser}[5]{
\ResultsFigStartTeaser
\multicolumn{2}{c}{(a) Reference} & \multicolumn{2}{c}{(b) Simplified Input} & \multicolumn{2}{c}{(c) RGBx Baseline} & \multicolumn{2}{c}{(d) Our Result} \\
\addlinespace[4pt]
\ResultsFigEndTeaser{#1}{#2}{#3}{#4}{#5}
}


\newcommand{\ResultsFigStartWithZoom}{
\setlength{\tabcolsep}{1pt}
\setlength{\h}{1.15in}
\begin{tabular}{cccccc}
}

\newcommand{\ResultsFigEndWithZoom}[6]{
\includegraphics[height=\h]{result_figs/#2_gt_box.png} & \includegraphics[height=\h]{result_figs/#2_ours_box.png} & \includegraphics[height=\h]{result_figs/#2_gt_zoom.png} & \includegraphics[height=\h]{result_figs/#2_MSAA_zoom.png} & \includegraphics[height=\h]{result_figs/#2_RGBx_zoom.png} & 
\includegraphics[height=\h]{result_figs/#2_ours_zoom.png} \tablegap
{\small {#1}} & & 
{\small {#3}} & 
{\small {#6}} & 
{\small {#5}} & 
{\small {#4}} 
\end{tabular}
}

\newcommand{\ResultsFigWithHeaderWithZoom}[6]{
\ResultsFigStartWithZoom
(a) Reference & (b) Our Result & (c) Reference & (d) Supersample & (e) RGBx & (f) Ours\\
\addlinespace[4pt]
\ResultsFigEndWithZoom{#1}{#2}{#3}{#4}{#5}{#6}
}

\newcommand{\ResultsFigNoHeaderWithZoom}[6]{
\ResultsFigStartWithZoom
\addlinespace[4pt]
\ResultsFigEndWithZoom{#1}{#2}{#3}{#4}{#5}{#6}
}

\newcommand{\ResultsFigEndWithCorrectedZoom}[6]{
\includegraphics[height=\h]{result_figs/#2_gt_box.png} & \includegraphics[height=\h]{result_figs/#2_ours_box.png} & \includegraphics[height=\h]{result_figs/#2_corrected_gt_zoom.png} & \includegraphics[height=\h]{result_figs/#2_corrected_MSAA_zoom.png} & \includegraphics[height=\h]{result_figs/#2_corrected_RGBx_zoom.png} & 
\includegraphics[height=\h]{result_figs/#2_corrected_ours_zoom.png} \tablegap
{\small {#1}} & & 
{\small {#3}} & 
{\small {#6}} & 
{\small {#5}} & 
{\small {#4}} 
\end{tabular}
}

\newcommand{\ResultsFigWithHeaderWithCorrectedZoom}[6]{
\ResultsFigStartWithZoom
(a) Reference & (b) Our Result & (c) Reference & (d) Supersample & (e) RGBx & (f) Ours\\
\addlinespace[4pt]
\ResultsFigEndWithCorrectedZoom{#1}{#2}{#3}{#4}{#5}{#6}
}

\newcommand{\ResultsFigNoHeaderWithCorrectedZoom}[6]{
\ResultsFigStartWithZoom
\addlinespace[4pt]
\ResultsFigEndWithCorrectedZoom{#1}{#2}{#3}{#4}{#5}{#6}
}


\newcommand{\ResultsFigEndInputWithZoom}[6]{
\includegraphics[height=\h]{result_figs/#2_gt_box.png} & \includegraphics[height=\h]{result_figs/#2_ours_box.png} & \includegraphics[height=\h]{result_figs/#2_gt_zoom.png} & \includegraphics[height=\h]{result_figs/#2_input_zoom.png} & \includegraphics[height=\h]{result_figs/#2_RGBx_zoom.png} & 
\includegraphics[height=\h]{result_figs/#2_ours_zoom.png} \tablegap
{\small {#1}} & & 
{\small {#3}} & 
{\small {#6}} & 
{\small {#5}} & 
{\small {#4}} 
\end{tabular}
}

\newcommand{\ResultsFigWithHeaderInputWithZoom}[6]{
\ResultsFigStartWithZoom
(a) Reference & (b) Our Result & (c) Reference & (d) Input & (e) RGBx & (f) Ours\\
\addlinespace[4pt]
\ResultsFigEndInputWithZoom{#1}{#2}{#3}{#4}{#5}{#6}
}

\newcommand{\ResultsFigNoHeaderInputWithZoom}[6]{
\ResultsFigStartWithZoom
\addlinespace[4pt]
\ResultsFigEndInputWithZoom{#1}{#2}{#3}{#4}{#5}{#6}
}


\newcommand{\ResultsFigStartWithoutZoom}{
\setlength{\tabcolsep}{1pt}
\setlength{\h}{1.15in}
\begin{tabular}{cccc}
}

\newcommand{\ResultsFigEndWithoutZoom}[5]{
\includegraphics[height=\h]{result_figs/#2_gt.png} & \includegraphics[height=\h]{result_figs/#2_input.png} & \includegraphics[height=\h]{result_figs/#2_RGBx.png} & \includegraphics[height=\h]{result_figs/#2_ours.png} \tablegap
{\small {#1}} & {\small {#5}} &
{\small {#4}} & 
{\small {#3}} 
\end{tabular}
}

\newcommand{\ResultsFigWithHeaderWithoutZoom}[5]{
\ResultsFigStartWithoutZoom
(a) Reference & (b) Simplified Input & (c) RGBx Baseline & (d) Our Result \\
\addlinespace[4pt]
\ResultsFigEndWithoutZoom{#1}{#2}{#3}{#4}{#5}
}

\newcommand{\ResultsFigNoHeaderWithoutZoom}[5]{
\ResultsFigStartWithoutZoom
\addlinespace[4pt]
\ResultsFigEndWithoutZoom{#1}{#2}{#3}{#4}{#5}
}


\newcommand{\ResultsFigStartThreeCol}{
\setlength{\tabcolsep}{1pt}
\setlength{\h}{1.15in}
\begin{tabular}{cc@{\hskip 4pt}cc@{\hskip 4pt}cc}
}

\newcommand{\ResultsFigEndThreeCol}[4]{
\includegraphics[height=\h]{result_figs/#2_gt_box.png} & \includegraphics[height=\h]{result_figs/#2_gt_zoom.png} & \includegraphics[height=\h]{result_figs/#2_RGBx_box.png} & \includegraphics[height=\h]{result_figs/#2_RGBx_zoom.png} & \includegraphics[height=\h]{result_figs/#2_ours_box.png} & \includegraphics[height=\h]{result_figs/#2_ours_zoom.png} \tablegap
\multicolumn{2}{c}{\small {#1}} &
\multicolumn{2}{c}{\small {#3}} &
\multicolumn{2}{c}{\small {#4}} 
\end{tabular}
}

\newcommand{\ResultsFigWithHeaderThreeCol}[4]{
\ResultsFigStartThreeCol
\multicolumn{2}{c}{(a) Reference} & \multicolumn{2}{c}{(b) RGBx Baseline} & \multicolumn{2}{c}{(c) Our Result} \\
\addlinespace[4pt]
\ResultsFigEndThreeCol{#1}{#2}{#3}{#4}
}

\newcommand{\ResultsFigNoHeaderThreeCol}[4]{
\ResultsFigStartThreeCol
\addlinespace[4pt]
\ResultsFigEndThreeCol{#1}{#2}{#3}{#4}
}

%% file: figtex/teaser.tex
\teaser{
\vspace{-2\baselineskip}    
\ResultsFigTeaser{\venice: 1x speedup, 0\% error}{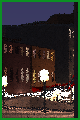}{1400x speedup, 7.4x error}{1300x speedup, 100\% error (baseline)}{1300x speedup, 70\% error}
  \caption{Learning to extrapolate from the partial computation of a procedural shader called \venice. The reference solution (a) results from a full computation at \changedi{1000} samples per pixel (\spp). A simplified version of the shader provides an approximate solution (b) using only \nspp{1} and less computation per sample (\changedi{72}\% of the original compute; \changedi{1400}x speedup overall). Our RGBx baseline method (c) learns to approximate the reference well for much of the image, based on only the RGB output of the simplified shader as well as a few hand-picked auxiliary features~-- but exhibits artifacts in the distance (obvious in the zooms boxed in green). This paper shows that difficult learning tasks like this can benefit from relying on not just the RGBx features but also the \emph{program trace} (a record of the intermediate values computed at every pixel, in this case that of the simplified shader) -- producing a more faithful approximation of the reference. Percent (\%) denotes mean perceptual error~\cite{zhang2018perceptual} relative to that of the RGBx baseline, averaged over the test set.}
\vspace{2ex}
\label{fig:teaser}
}

%% file: 0-abstract.tex
\begin{abstract}
Deep learning for image processing typically treats input imagery as pixels in some color space. 
This paper proposes instead to learn from program traces of procedural fragment shaders -- programs that generate images. 
At each pixel, we collect the intermediate values computed at program execution, and these data form the input to the learned model. 
We investigate this learning task for a variety of applications: 
our model can learn to predict a low-noise output image from shader programs that exhibit sampling noise; 
this model can also learn from a simplified shader program that approximates the reference solution with less computation,
as well as learn the output of postprocessing filters like defocus blur and edge-aware sharpening.
Finally we show that the idea of learning from program traces can even be applied to non-imagery simulations of flocks of boids.
Our experiments on a variety of shaders show quantitatively and qualitatively that models learned from program traces outperform baseline
models learned from RGB color augmented with hand-picked shader-specific features like normals, depth, and diffuse and specular color.
%
%
%
\changedh{We also conduct a series of analyses that show certain features within the trace are more important, and even learning from a small subset of the trace outperforms the baselines.}

\ccsdesc[500]{Computing methodologies~Neural networks}
\ccsdesc[500]{Software and its engineering~Compilers}
\ccsdesc[500]{Computing methodologies~Computer graphics}

\printccsdesc

\end{abstract}


%% file: figtex/summary_bar.tex
\begin{figure}
    \centering
    \hspace*{-0.1in}\includegraphics[width=3.5in]{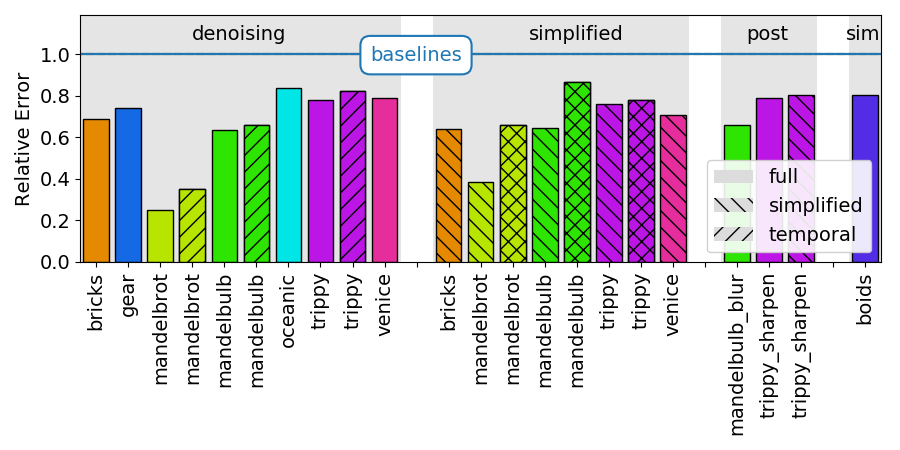}
\vspace*{-1.75\baselineskip}
    \caption{\changedj{Four-application summary: {\textbf{denoising}}, reconstruction from {\textbf{simplified}} shaders, learned {\textbf{post}}-processing effects and {\textbf{sim}}ulation.
The vertical axis shows (in every example)} improved perceptual error compared to each applications' strongest baseline:
RGBx for denoising, simplified, and post; and I/O for simulation (\sect{sec:eval}). 
Hatching directions denote use of simplified shaders and/or temporally coherent models.
}
    \label{fig:summary_bar}
\vspace*{-1.0\baselineskip}
\end{figure}

%% file: figtex/arch.tex
\begin{figure}[b!]
    \centering
    \includegraphics[width=3.25in]{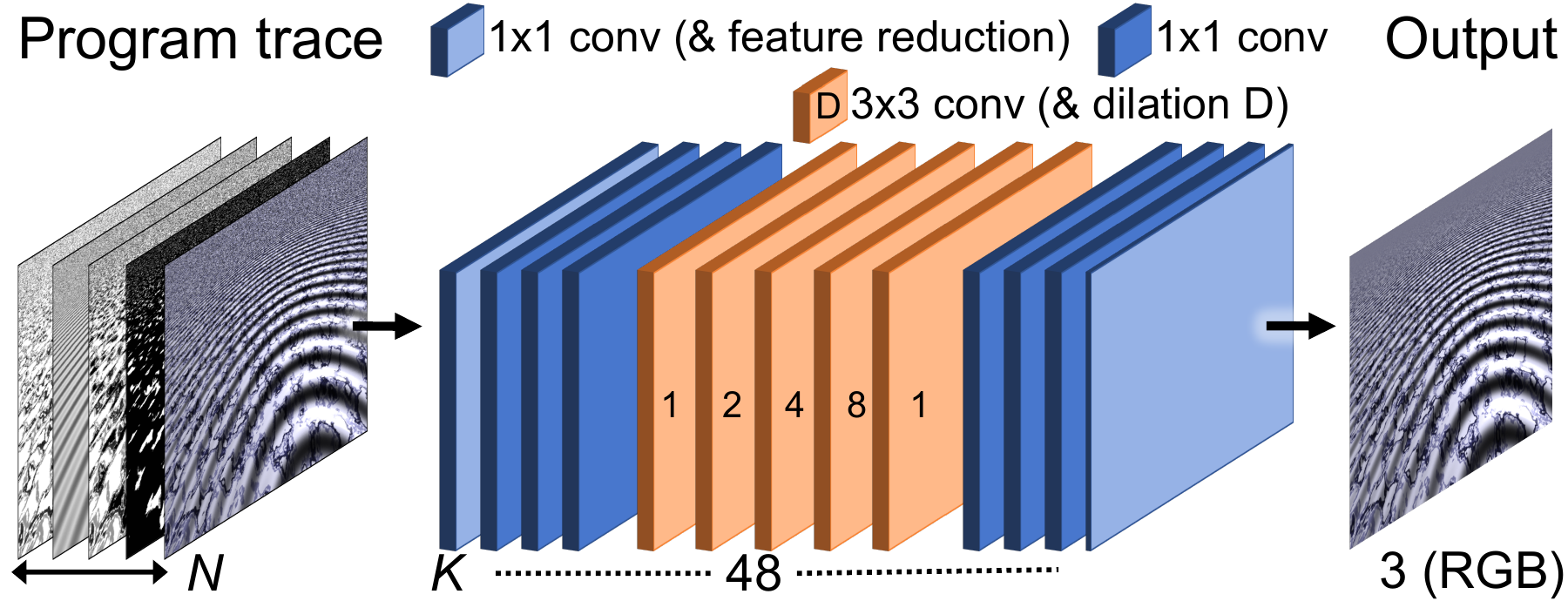}
    \caption{Network architecture used in our experiments. The input from the program trace has \tracelen\ channels. The output layer has three channels for color images. \changedi{The first feature reduction layer has \reducedim\ channels. We use $\reducedim = 48$ in our method. When training the baseline method, \reducedim\ will be increased to a larger value to match the total number of trainable weights to be the same as training with the program trace at the maximum length.} All \changedi{other} intermediate layers have 48 channels. The input feature maps are first analyzed by four 1x1 convolutional layers, followed by five 3x3 convolutional layers with dilation rates of ${1, 2, 4, 8, 1}$ respectively. Finally, four additional 1x1 convolutional layers are applied and output a three channel image. Note that the first and last convolutional blocks indicated in lighter blue each reduce the number of channels (from \tracelen\ to 48, and from 48 to 3, respectively).}
    \label{fig:architecture}
\end{figure}

%% file: figtex/result_fig_denoising.tex
\begin{figure*}

\ResultsFigWithHeaderWithZoom{\bricks}{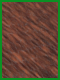}{0\%/1x}{68\%/930x}{100\%/960x}{1\,SPP, 21x/1000x}

\ResultsFigNoHeaderWithZoom{\mandelbrot}{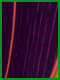}{0\%/1x}{24\%/740x}{100\%/780x}{2\,SPP, 15x/520x}

\ResultsFigNoHeaderWithZoom{\oceanic}{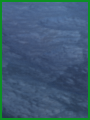}{0\%/1x}{83\%/990x}{100\%/1000x}{1\,SPP, 11x/1000x}
    
\ResultsFigNoHeaderWithZoom{\primitives}{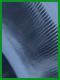}{0\%/1x}{74\%/960x}{100\%/980x}{1\,SPP, 17x/1000x}

\vspace{-2ex}
\caption{Learning to reduce sampling noise in procedural shaders.
The reference low-noise solution (a) relies on 1000 samples per pixel (\spp). Our method (b) approximates the reference well at only \nspp{1}. Zooming into the region boxed in green (c, f) reveals approximation error, which compares favorably with the two baselines: (d) supersampling where the number of samples is chosen to have comparable run-time as ours, and (e) RGBx.
Error and speedups are reported as in \fig{fig:teaser}.
%
Our method better covers both the orientation and high-frequency detail than the baselines. 
\bricks\ and \mandelbrot\ are gamma corrected to emphasize visual differences.
}
\vspace{-4ex}
\label{fig:result_denoising}
\end{figure*}

%% file: figtex/error_table_main.tex
\setlength{\tabcolsep}{2.5pt}
\begin{table}[t]

\centering

\caption{Error statistics for applications in \sect{sec:eval} and \ref{sec:multiple_shaders}. Errors reported as LPIPS \cite{zhang2018perceptual} / SSIM / PSNR. 
}

\begin{tabular}{c|l|c@{ / }c@{ / }cc@{ / }c@{ / }c}
        \hline
         & Shader & \multicolumn{3}{c}{RGBx} & \multicolumn{3}{c}{Ours} \\ \thickhline
        
        {\multirow{7}{*}{\rotatebox[origin=c]{90}{Denoising}}} & \bricks & 0.0141 & 0.981 & 36.68 & \textbf{0.0097} & \textbf{0.987} & \textbf{38.29} \\ 
     \cline{2-8} 
                         & \gear & 0.0173 & 0.986 & 38.90 & \textbf{0.0127} & \textbf{0.988} & \textbf{39.86} \\ 
     \cline{2-8} 
                         & \mandelbrot & 0.0235 & 0.973 & 36.07 & \textbf{0.0059} & \textbf{0.986} & \textbf{38.55} \\ 
     \cline{2-8} 
                         & \mandelbulb & 0.0185 & 0.962 & 32.14 & \textbf{0.0118} & \textbf{0.975} & \textbf{34.18} \\ 
     \cline{2-8} 
                         & \oceanic & 0.0403 & 0.961 & 33.69 & \textbf{0.0339} & \textbf{0.966} & \textbf{34.51} \\ 
     \cline{2-8} 
                         & \trippy & 0.0696 & 0.856 & 26.30 & \textbf{0.0543} & \textbf{0.886} & \textbf{27.27} \\ 
     \cline{2-8} 
                         & \venice & 0.0309 & 0.965 & 32.76 & \textbf{0.0242} & \textbf{0.973} & \textbf{33.83} \\ 
     \thickhline 
        {\multirow{5}{*}{\rotatebox[origin=c]{90}{Simplified}}} & \bricks & 0.0624 & 0.922 & 25.57 & \textbf{0.0398} & \textbf{0.936} & \textbf{29.96} \\ 
     \cline{2-8} 
                         & \mandelbrot & 0.1111 & 0.826 & 27.04 & \textbf{0.0430} & \textbf{0.949} & \textbf{31.15} \\ 
     \cline{2-8} 
                         & \mandelbulb & 0.0932 & 0.812 & 25.22 & \textbf{0.0600} & \textbf{0.856} & \textbf{26.85} \\ 
     \cline{2-8} 
                         & \trippy & 0.2412 & 0.520 & 18.55 & \textbf{0.1824} & \textbf{0.629} & \textbf{20.95} \\ 
     \cline{2-8} 
                         & \venice & 0.0404 & 0.957 & 31.50 & \textbf{0.0285} & \textbf{0.965} & \textbf{32.72} \\ 
     \thickhline 
        {\multirow{3}{*}{\rotatebox[origin=c]{90}{Post}}} & blur & 0.0126 & 0.978 & 35.81 & \textbf{0.0082} & \textbf{0.985} & \textbf{37.62} \\ 
     \cline{2-8} 
                         & Sharpen & 0.0881 & 0.833 & 24.26 & \textbf{0.0693} & \textbf{0.868} & \textbf{25.31} \\ 
     \cline{2-8} 
                         & Simp Sharpen & 0.2675 & 0.477 & 17.20 & \textbf{0.2154} & \textbf{0.587} & \textbf{19.35} \\ 
     \thickhline 
     {\multirow{4}{*}{\rotatebox[origin=c]{90}{Shared}}} & \gear & 0.0251 & 0.984 & 38.16 & \textbf{0.0173} & \textbf{0.986} & \textbf{38.66} \\ 
     \cline{2-8} 
                         & \mandelbrot & 0.0546 & 0.801 & 28.02 & \textbf{0.0165} & \textbf{0.933} & \textbf{32.60} \\ 
     \cline{2-8} 
                         & \mandelbulb & 0.0423 & 0.861 & 28.02 & \textbf{0.0298} & \textbf{0.906} & \textbf{30.19} \\ 
     \cline{2-8} 
                         & \trippy & 0.1048 & 0.815 & 19.99 & \textbf{0.0755} & \textbf{0.857} & \textbf{23.88} \\ 
     \thickhline 
    \end{tabular}

\label{tab:err_main}

\vspace{-2ex}
\end{table}

%% file: figtex/result_fig_simplified.tex
\begin{figure*}

    \ResultsFigWithHeaderInputWithZoom{\mandelbrot}{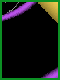}{0\%/1x}{38\%/1300x}{100\%/1500x}{4.4x/2400x}
    
    \ResultsFigNoHeaderInputWithZoom{\mandelbulb}{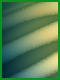}{0\%/1x}{64\%/920x}{100\%/1100x}{2.9x/1800x}
    
    \ResultsFigNoHeaderInputWithZoom{\trippy}{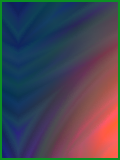}{0\%/1x}{75\%/270x}{100\%/340x}{2.8x/1600x}
    
\vspace{-2ex}
\caption{\changedi{Learning from simplified shaders \mandelbrot, \mandelbulb\ and \trippy. Errors and speedups are reported as in \fig{fig:teaser}. In \mandelbrot\ our method better reconstructs missing regions due to oversimplification in the input. In \mandelbulb\ our method better recovers the orientation of the texture. In \trippy\ ours correctly recovers the color.}}
\label{fig:result_simplified}

\vspace*{-1.0\baselineskip}

\end{figure*}

%% file: figtex/result_fig_post_processing.tex
\begin{figure*}

   \ResultsFigWithHeaderThreeCol{\mandelbulb\ blur: 0\%/1x}{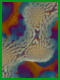}{100\%/700x (baseline)}{65\%/590x}
    
    \ResultsFigNoHeaderThreeCol{\trippy\ simplified\ sharpen: 0\%/1x}{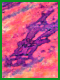}{100\%/340x (baseline)}{80\%/270x}
    
\vspace{-2ex}
\caption{Learning postprocessing effects. The reference solution (a) shows the result of a postprocessing filter applied to a low-noise shader rendering sampled at \nspp{1000}. Both RGBx baseline (b) and our method (c) approximates the reference at \nspp{1}. Our method recovers more faithfully the thin structure in \mandelbulb\ and the color pattern in \trippy. We report relative perceptual error \changedi{and speedup} as in \fig{fig:teaser}. \changedi{\mandelbulb\ is gamma corrected so it can be viewed comfortably on darker displays.}}
\label{fig:result_fig_post_processing}
\end{figure*}

%% file: figtex/simulation_boids.tex
\begin{figure}[b!]
\vspace{-2pt}
\centering
\setlength{\h}{1.17in}
    \centering
    \begin{tabular}{cc}
\includegraphics[height=\h]{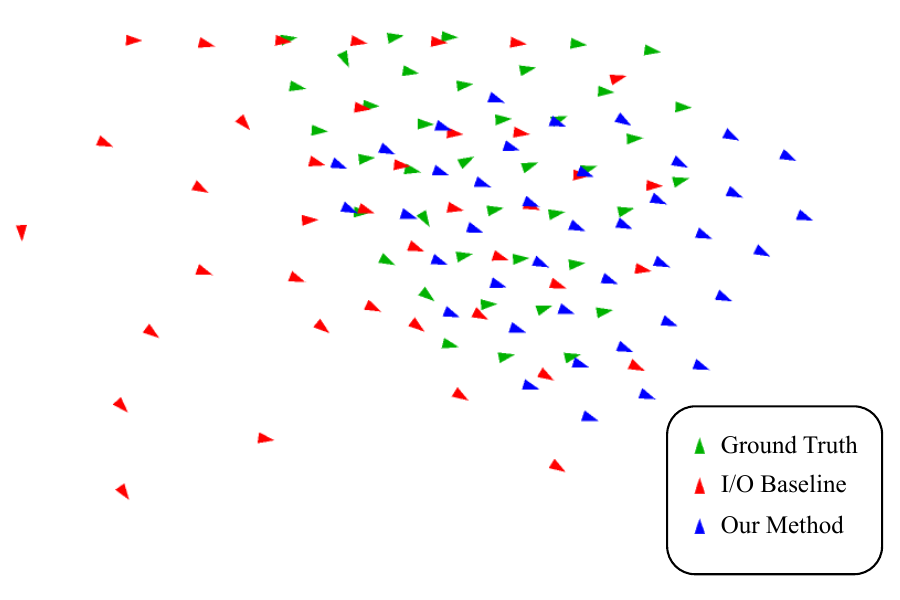} & \includegraphics[height=\h]{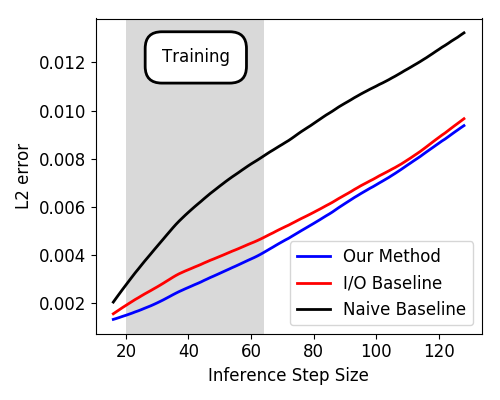} \\
(a) Visualization & (b) Error Analysis \\
\end{tabular}
\caption{(a) Visualization for flocks of boids. Both I/O baseline (red) and our method (blue) starts from the same initial state and have taken \changedi{80} inference steps with step size 20. Our mean position as well as flocking behavior is more faithful to the ground truth (green). (b) We plot average error as a function of step size, where training ranges from step size 20 to 64 (gray). Ours consistently outperforms both I/O and a more naive baseline, in the training range and beyond it.}

    \label{fig:simulation_boids}
\end{figure}

%% file: figtex/trace_subsample_comparison.tex
\begin{figure}
    \centering
    \includegraphics[width=3.3in]{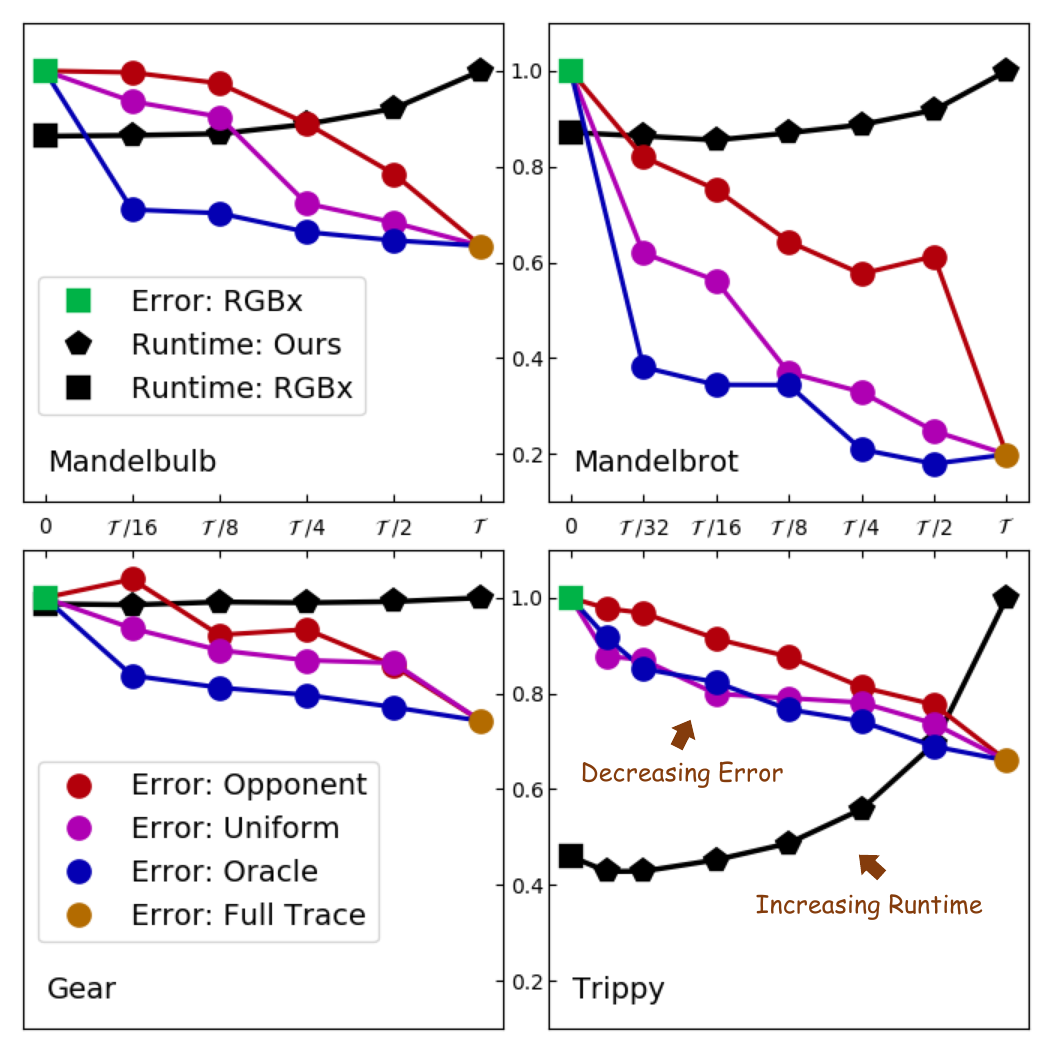}
    \caption{\changedi{Error vs. Time trade-off for Opponent, Uniform, and Oracle subsampling strategies with varying trace length. For each shader with \trace\ program traces, we subsample the trace such that the actual input trace length \tracelen\ is equal to \trace/2, \trace/4, \trace/8, etc., and the x-axis shows each model's relative trace ratio compared to the full program trace. Circle shows perceptual error relative to the RGBx baseline (green square) for Opponent (red), Uniform (purple), Oracle (blue), and Full Trace (gold). Black pentagon shows relative inference time (including shader program runtime and network inference) for each \tracelen\ relative to the runtime of the Full Trace model. Similarly black square shows relative inference time for the RGBx baseline. Note that the x-axis is on a log scale in relative trace length compared to the Full Trace model, therefore although the relative error plot appears linear as \tracelen\ increases, the actual performance improvement is faster at the beginning of the plot: adding only a few traces quickly improves performance.}}

    \label{fig:trace_subsample_comparison}
\end{figure}

%% file: figtex/error_table_temporal.tex
\setlength{\tabcolsep}{2pt}
\begin{table}[!bt]

\centering

\caption{Error statistic for learning temporally coherence sequences.
Metrics reported are similar as in \tbl{tab:err_main}.
The temporal application is trained both on shaders with full computation and simplified shaders with partial computation (simp). For each experiment, we generate a 30 frames sequence and compute the error with respect to ground truth using the last frame. The reported numbers are averaged across 30 different sequences.
}

\begin{tabular}{l|c@{ / }c@{ / }cc@{ / }c@{ / }c}
        \hline
        Shader & \multicolumn{3}{c}{RGBx} & \multicolumn{3}{c}{Ours} \\ \thickhline
        
        \mandelbrot\  & 0.0104 & 0.980 & 37.16 & \textbf{0.0037} & \textbf{0.989} & \textbf{39.75} \\ 
     \cline{1-7} 
                         \mandelbrot\ simp  & 0.1049 & 0.898 & 27.51 & \textbf{0.0693} & \textbf{0.929} & \textbf{28.93} \\ 
     \cline{1-7}
                         \mandelbulb\  & 0.0213 & 0.959 & 32.06 & \textbf{0.0140} & \textbf{0.971} & \textbf{33.56} \\ 
     \cline{1-7}
                         \mandelbulb\ simp  & 0.1194 & 0.780 & 23.49 & \textbf{0.1035} & \textbf{0.788} & \textbf{24.25} \\ 
     \cline{1-7}
                         \trippy\  & 0.0665 & 0.864 & 26.61 & \textbf{0.0546} & \textbf{0.884} & \textbf{27.17} \\ 
     \cline{1-7}
                         \trippy\ simp  & 0.2295 & 0.563 & 19.10 & \textbf{0.1788} & \textbf{0.637} & \textbf{21.10} \\ 
     \thickhline 
    \end{tabular}

\vspace{2ex}

\label{tab:err_temporal}
\end{table}

%% file: figtex/result_fig_temporal.tex
\begin{figure}[b!]
\centering
\setlength{\h}{1.6in}
    \centering
    \begin{tabular}{cc}
(a) RGBx Baseline & (b) Our result \\
\includegraphics[width=\h]{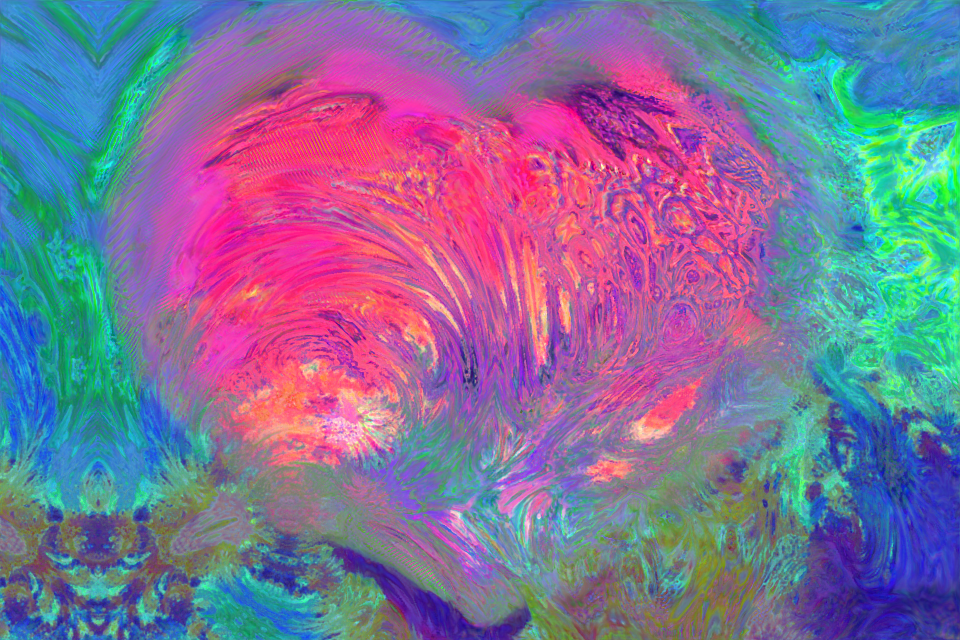} & \includegraphics[width=\h]{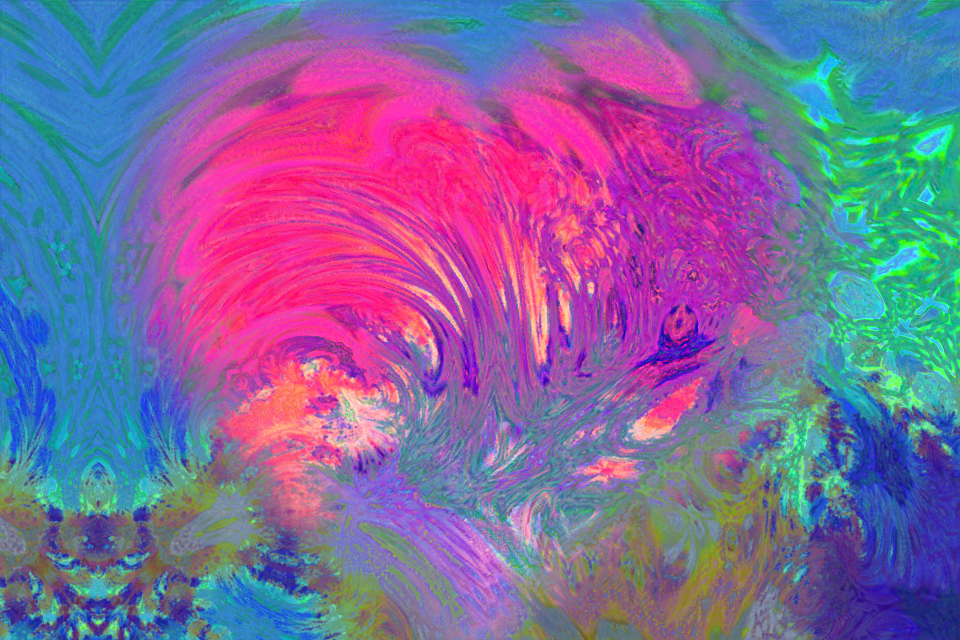} \\
\small{\trippy: 100\%} & \small{\changedi{77}\%} \\
\end{tabular}
    \caption{Learning temporally coherent sequences for \trippy\ with the same ground truth as in \fig{fig:result_simplified}. We report relative perceptual error compared to the RGBx baseline. Both RGBx (a) and ours (b) are the 90th frame of a synthesized temporally coherent sequence. Note how our method generalizes well to long sequences whereas the RGBx baseline presents obvious artifacts \changedi{such as color residual from previous frames near the silhouette of the heart.}}
    \label{fig:result_temporal}
\end{figure}

%% file: figtex/simulation_fluid.tex
\begin{figure}[b!]
\centering
\setlength{\h}{1.05in}
    \centering
    \begin{tabular}{ccc}
(a) Reference & (b) I/O Baseline & (c) Our Method \\
\includegraphics[width=\h]{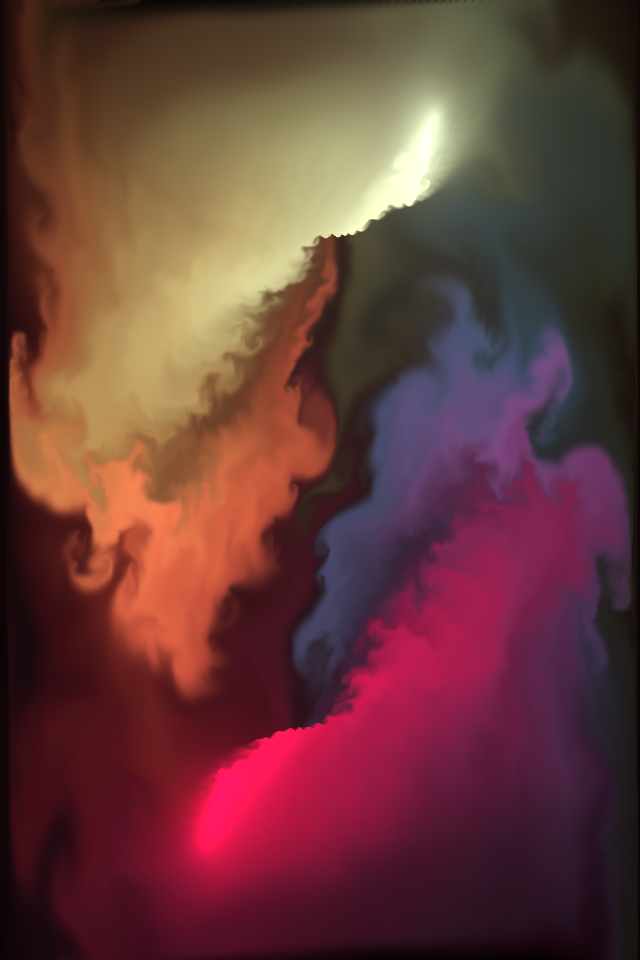} & \includegraphics[width=\h]{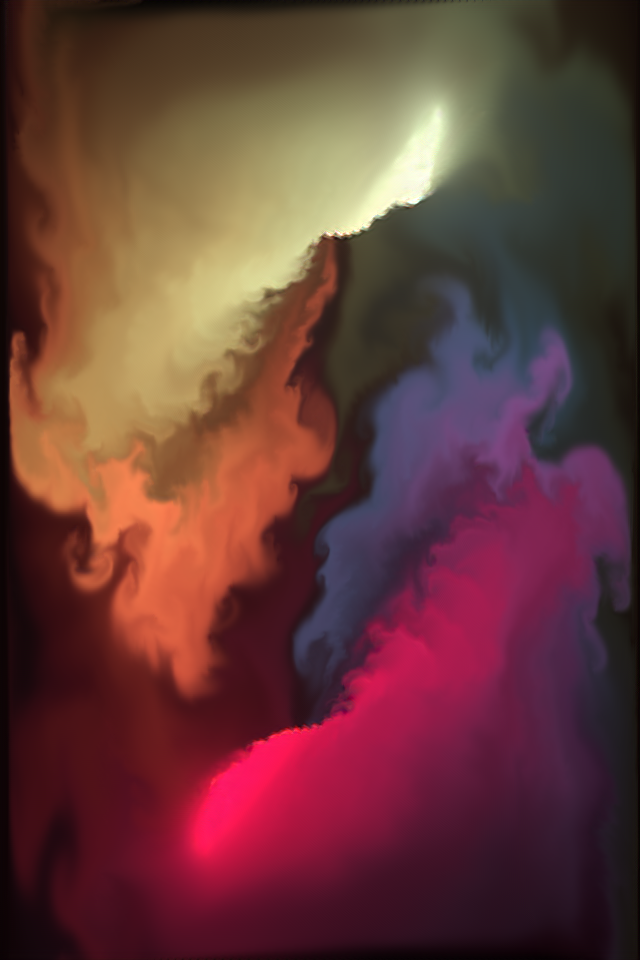} & \includegraphics[width=\h]{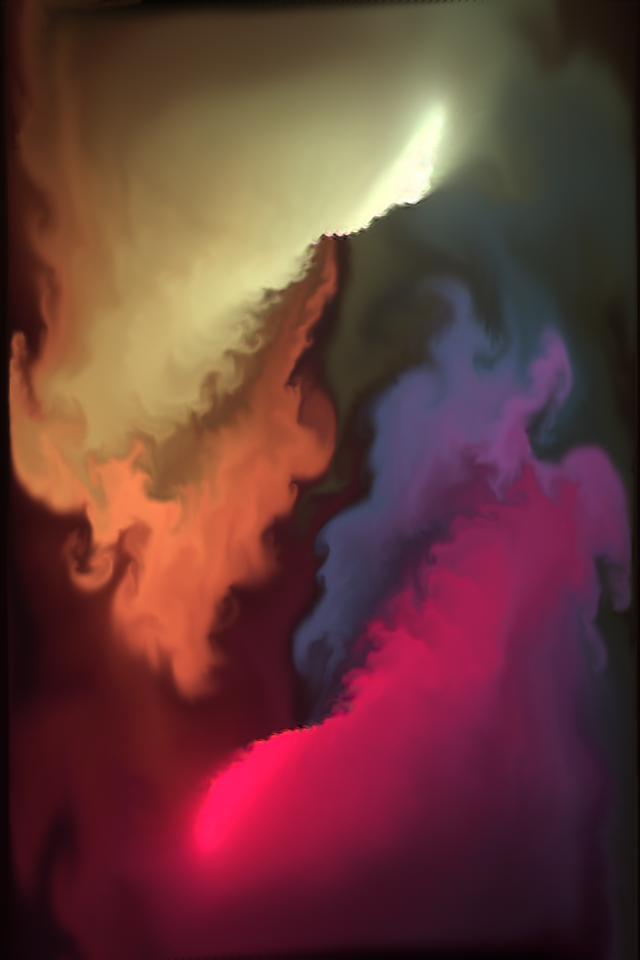} \\
\small{0\%} & \small{100\%} & \small{96\%} \\
\end{tabular}
\caption{An example of fluid simulation where our method (c) gives a very similar result as the I/O baseline (b). This indicates our method may not be advantageous for simple learning tasks where the baseline is already good enough to reconstruct the reference (a). The relative perceptual error compared to the I/O baseline is reported below each image.}
    \label{fig:simulation_fluid}
\end{figure}

%% file: figtex/error_table_denoising.tex
\setlength{\tabcolsep}{10pt}
\begin{table*}[]
\centering
\caption{Error statistics for denoising (\sect{sec:denoising}). We report LPIPS perceptual error \cite{zhang2018perceptual}, SSIM and PSNR. The numbers reported are averaged across the entire test dataset. 
The error metric in the first two columns (RGBx and ours) are already reported in the main paper \tbl{tab:err_main}, we include them here for a clear comparison with Supersampling.
For LPIPS, we report the absolute error for the RGBx baseline, and for other methods their errors relative to that of the RGBx baseline (\% or $\times$).}

\begin{tabular}{l|c@{ / }c@{ / }cc@{ / }c@{ / }cc@{(}r@{) / }c@{ / }c}
    \hline
    
    Shader & \multicolumn{3}{c}{RGBx} & \multicolumn{3}{c}{Ours} & \multicolumn{4}{c}{Supersampling} \\ \hline
    
            \bricks & 0.0141 & 0.981 & 36.68 & \textbf{0.0097(68\%)} & \textbf{0.987} & \textbf{38.29} & 0.2985 & 21$\times$ & 0.839 & 24.14 \\ \hline
            \gear & 0.0173 & 0.986 & 38.90 & \textbf{0.0127(73\%)} & \textbf{0.988} & \textbf{39.86} & 0.2935 & 17$\times$ & 0.880 & 24.71 \\ \hline
            \mandelbrot & 0.0235 & 0.973 & 36.07 & \textbf{0.0059(24\%)} & \textbf{0.986} & \textbf{38.55} & 0.3502 & 15$\times$ & 0.746 & 23.46 \\ \hline
            \mandelbulb & 0.0185 & 0.962 & 32.14 & \textbf{0.0118(63\%)} & \textbf{0.975} & \textbf{34.18} & 0.1580 & 8.5$\times$ & 0.895 & 23.98 \\ \hline
            \oceanic & 0.0403 & 0.961 & 33.69 & \textbf{0.0339(84\%)} & \textbf{0.966} & \textbf{34.51} & 0.4314 & 11$\times$ & 0.780 & 23.81 \\ \hline
            \trippy & 0.0696 & 0.856 & 26.30 & \textbf{0.0543(77\%)} & \textbf{0.886} & \textbf{27.27} & 0.2178 & 3.1$\times$ & 0.767 & 22.42 \\ \hline
            \venice & 0.0309 & 0.965 & 32.76 & \textbf{0.0242(78\%)} & \textbf{0.973} & \textbf{33.83} & 0.2893 & 9.4$\times$ & 0.853 & 23.73 \\ \hline

\end{tabular}

\label{tab:error_denoising}
\end{table*}

%% file: main.bbl
\newcommand{\etalchar}[1]{$^{#1}$}
\begin{thebibliography}{\uppercase{SDMHR11}}

\bibitem[AMHH08]{akenine2008real}
\textsc{Akenine-M{\"o}ller T., Haines E., Hoffman N.}:
\newblock \emph{Real-time rendering}.
\newblock CRC Press, 2008.

\bibitem[BA83]{burt1983laplacian}
\textsc{Burt P., Adelson E.}:
\newblock The laplacian pyramid as a compact image code.
\newblock \emph{IEEE Transactions on communications 31}, 4 (1983), 532--540.

\bibitem[BJB{\etalchar{*}}19]{mit-saliency-benchmark}
\textsc{Bylinskii Z., Judd T., Borji A., Itti L., Durand F., Oliva A., Torralba
  A.}:
\newblock Mit saliency benchmark, 2019.

\bibitem[CBSC16]{MLNet_Saliency}
\textsc{Cornia M., Baraldi L., Serra G., Cucchiara R.}:
\newblock A deep multi-level network for saliency prediction.
\newblock In \emph{2016 23rd International Conference on Pattern Recognition
  (ICPR)} (2016), pp.~3488--3493.

\bibitem[CKS{\etalchar{*}}17]{Chaitanya}
\textsc{Chaitanya C. R.~A., Kaplanyan A.~S., Schied C., Salvi M., Lefohn A.,
  Nowrouzezahrai D., Aila T.}:
\newblock Interactive reconstruction of monte carlo image sequences using a
  recurrent denoising autoencoder.
\newblock \emph{ACM Trans. Graph. 36}, 4 (July 2017), 98:1--98:12.

\bibitem[CLS19]{chen2018executionguided}
\textsc{Chen X., Liu C., Song D.}:
\newblock Execution-guided neural program synthesis.
\newblock In \emph{International Conference on Learning Representations}
  (2019).

\bibitem[CSS18]{DBLP:journals/corr/abs-1801-02318}
\textsc{Chen L., Sultana S., Sahita R.}:
\newblock Henet: {A} deep learning approach on intel{\textregistered} processor
  trace for effective exploit detection.
\newblock \emph{CoRR abs/1801.02318} (2018).

\bibitem[CXK17]{Chen_2017_ICCV}
\textsc{Chen Q., Xu J., Koltun V.}:
\newblock Fast image processing with fully-convolutional networks.
\newblock In \emph{The IEEE International Conference on Computer Vision (ICCV)}
  (Oct 2017).

\bibitem[DBLW15]{dorn2015towards}
\textsc{Dorn J., Barnes C., Lawrence J., Weimer W.}:
\newblock Towards automatic band-limited procedural shaders.
\newblock In \emph{Computer Graphics Forum} (2015), vol.~34, Wiley Online
  Library, pp.~77--87.

\bibitem[FC18]{frankle2018lottery}
\textsc{Frankle J., Carbin M.}:
\newblock The lottery ticket hypothesis: Finding sparse, trainable neural
  networks.
\newblock \emph{arXiv preprint arXiv:1803.03635} (2018).

\bibitem[FH93]{feiler1993software}
\textsc{Feiler P.~H., Humphrey W.~S.}:
\newblock Software process development and enactment: Concepts and definitions.
\newblock In \emph{Software Process, 1993. Continuous Software Process
  Improvement, Second International Conference on the} (1993), IEEE,
  pp.~28--40.

\bibitem[GCB{\etalchar{*}}17]{gharbi2017deep}
\textsc{Gharbi M., Chen J., Barron J.~T., Hasinoff S.~W., Durand F.}:
\newblock Deep bilateral learning for real-time image enhancement.
\newblock \emph{ACM Transactions on Graphics (TOG) 36}, 4 (2017), 118.

\bibitem[GKB{\etalchar{*}}18]{DBLP:journals/corr/abs-1804-01118}
\textsc{Ganin Y., Kulkarni T., Babuschkin I., Eslami S. M.~A., Vinyals O.}:
\newblock Synthesizing programs for images using reinforced adversarial
  learning.
\newblock \emph{CoRR abs/1804.01118} (2018).

\bibitem[GLA{\etalchar{*}}19]{gharbi2019sample}
\textsc{Gharbi M., Li T.-M., Aittala M., Lehtinen J., Durand F.}:
\newblock Sample-based monte carlo denoising using a kernel-splatting network.
\newblock \emph{ACM Transactions on Graphics (TOG) 38}, 4 (2019), 1--12.

\bibitem[Goo17]{DBLP:journals/corr/Goodfellow17}
\textsc{Goodfellow I.}:
\newblock {NIPS} 2016 tutorial: Generative adversarial networks.
\newblock \emph{CoRR abs/1701.00160} (2017).

\bibitem[HFTF15]{he2015system}
\textsc{He Y., Foley T., Tatarchuk N., Fatahalian K.}:
\newblock A system for rapid, automatic shader level-of-detail.
\newblock \emph{ACM Transactions on Graphics (TOG) 34}, 6 (2015), 1--12.

\bibitem[HPG{\etalchar{*}}19]{DBLP:journals/corr/abs-1905-04077}
\textsc{Hahn C., Phan T., Gabor T., Belzner L., Linnhoff{-}Popien C.}:
\newblock Emergent escape-based flocking behavior using multi-agent
  reinforcement learning.
\newblock \emph{CoRR abs/1905.04077} (2019).

\bibitem[IZZE17]{isola2017image}
\textsc{Isola P., Zhu J.-Y., Zhou T., Efros A.~A.}:
\newblock Image-to-image translation with conditional adversarial networks.
\newblock In \emph{Computer Vision and Pattern Recognition (CVPR), 2017 IEEE
  Conference on} (2017), IEEE, pp.~5967--5976.

\bibitem[JMKO20]{9218613}
\textsc{{Jeppu} N.~Y., {Melham} T., {Kroening} D., {O’Leary} J.}:
\newblock Learning concise models from long execution traces.
\newblock In \emph{2020 57th ACM/IEEE Design Automation Conference (DAC)}
  (2020), pp.~1--6.

\bibitem[KWGB17]{kummerer2017understanding}
\textsc{Kummerer M., Wallis T.~S., Gatys L.~A., Bethge M.}:
\newblock Understanding low-and high-level contributions to fixation
  prediction.
\newblock In \emph{Proceedings of the IEEE International Conference on Computer
  Vision} (2017), pp.~4789--4798.

\bibitem[KZP{\etalchar{*}}20]{Kim2020_GameGan}
\textsc{Kim S.~W., Zhou Y., Philion J., Torralba A., Fidler S.}:
\newblock {Learning to Simulate Dynamic Environments with GameGAN}.
\newblock In \emph{IEEE Conference on Computer Vision and Pattern Recognition
  (CVPR)} (Jun. 2020).

\bibitem[Lar93]{larus1993efficient}
\textsc{Larus J.~R.}:
\newblock Efficient program tracing.
\newblock \emph{Computer 26}, 5 (1993), 52--61.

\bibitem[LGA{\etalchar{*}}18]{li2018differentiable}
\textsc{Li T.-M., Gharbi M., Adams A., Durand F., Ragan-Kelley J.}:
\newblock Differentiable programming for image processing and deep learning in
  halide.
\newblock \emph{ACM Transactions on Graphics (TOG) 37}, 4 (2018), 139.

\bibitem[LHAY17]{li2017joint}
\textsc{Li Y., Huang J.-B., Ahuja N., Yang M.-H.}:
\newblock Joint image filtering with deep convolutional networks.
\newblock \emph{arXiv preprint arXiv:1710.04200} (2017).

\bibitem[LKD{\etalchar{*}}16]{DBLP:journals/corr/LiKDSG16}
\textsc{Li H., Kadav A., Durdanovic I., Samet H., Graf H.~P.}:
\newblock Pruning filters for efficient convnets.
\newblock \emph{CoRR abs/1608.08710} (2016).

\bibitem[LNLB16]{lou2016image}
\textsc{Lou L., Nguyen P., Lawrence J., Barnes C.}:
\newblock Image perforation: Automatically accelerating image pipelines by
  intelligently skipping samples.
\newblock \emph{ACM Transactions on Graphics (TOG) 35}, 5 (2016), 153.

\bibitem[LWL17]{Luo_2017_ICCV}
\textsc{Luo J.-H., Wu J., Lin W.}:
\newblock Thinet: A filter level pruning method for deep neural network
  compression.
\newblock In \emph{The IEEE International Conference on Computer Vision (ICCV)}
  (Oct 2017).

\bibitem[MMT{\etalchar{*}}19]{DBLP:journals/corr/abs-1906-10771}
\textsc{Molchanov P., Mallya A., Tyree S., Frosio I., Kautz J.}:
\newblock Importance estimation for neural network pruning.
\newblock \emph{CoRR abs/1906.10771} (2019).

\bibitem[MST{\etalchar{*}}20]{mildenhall2020nerf}
\textsc{Mildenhall B., Srinivasan P.~P., Tancik M., Barron J.~T., Ramamoorthi
  R., Ng R.}:
\newblock Nerf: Representing scenes as neural radiance fields for view
  synthesis.
\newblock \emph{arXiv preprint arXiv:2003.08934} (2020).

\bibitem[MTK{\etalchar{*}}16]{DBLP:journals/corr/MolchanovTKAK16}
\textsc{Molchanov P., Tyree S., Karras T., Aila T., Kautz J.}:
\newblock Pruning convolutional neural networks for resource efficient transfer
  learning.
\newblock \emph{CoRR abs/1611.06440} (2016).

\bibitem[NAM{\etalchar{*}}17]{Nalbach2017b}
\textsc{Nalbach O., Arabadzhiyska E., Mehta D., Seidel H.-P., Ritschel T.}:
\newblock Deep shading: Convolutional neural networks for screen-space shading.

\bibitem[Per01]{perlin2001noise}
\textsc{Perlin K.}:
\newblock Noise hardware.
\newblock \emph{Real-Time Shading SIGGRAPH Course Notes} (2001).

\bibitem[PHK11]{Paris:2011:LLF:1964921.1964963}
\textsc{Paris S., Hasinoff S.~W., Kautz J.}:
\newblock Local laplacian filters: Edge-aware image processing with a laplacian
  pyramid.
\newblock In \emph{ACM SIGGRAPH 2011 Papers} (2011), SIGGRAPH '11,
  pp.~68:1--68:12.

\bibitem[Rey87]{reynolds:1987:CG}
\textsc{Reynolds C.~W.}:
\newblock Flocks, herds, and schools: {A} distributed behavioral model.
\newblock \emph{SIGGRAPH Computer Graphics 21}, 4 (July 1987), 25--34.

\bibitem[RF16]{Reed2016NeuralP}
\textsc{Reed S., Freitas N.~D.}:
\newblock Neural programmer-interpreters.
\newblock \emph{CoRR abs/1511.06279} (2016).

\bibitem[Rok93]{rokita1993fast}
\textsc{Rokita P.}:
\newblock Fast generation of depth of field effects in computer graphics.
\newblock \emph{Computers \& Graphics 17}, 5 (1993), 593--595.

\bibitem[RTHG16]{NIPS2016_40008b9a}
\textsc{Ritchie D., Thomas A., Hanrahan P., Goodman N.}:
\newblock Neurally-guided procedural models: Amortized inference for procedural
  graphics programs using neural networks.
\newblock In \emph{Advances in Neural Information Processing Systems} (2016),
  Lee D., Sugiyama M., Luxburg U., Guyon I., Garnett R., (Eds.), vol.~29,
  Curran Associates, Inc., pp.~622--630.

\bibitem[SaMWL11]{genprog_simpl}
\textsc{Sitthi-amorn P., Modly N., Weimer W., Lawrence J.}:
\newblock Genetic programming for shader simplification.
\newblock \emph{ACM Trans. Graph. 30} (12 2011), 152.

\bibitem[SDMHR11]{sidiroglou2011managing}
\textsc{Sidiroglou-Douskos S., Misailovic S., Hoffmann H., Rinard M.}:
\newblock Managing performance vs. accuracy trade-offs with loop perforation.
\newblock In \emph{Proceedings of the 19th ACM SIGSOFT symposium and the 13th
  European conference on Foundations of software engineering} (2011), ACM,
  pp.~124--134.

\bibitem[STH{\etalchar{*}}19]{sitzmann2019deepvoxels}
\textsc{Sitzmann V., Thies J., Heide F., Nie{\ss}ner M., Wetzstein G.,
  Zollh{\"o}fer M.}:
\newblock Deepvoxels: Learning persistent 3d feature embeddings.
\newblock In \emph{Proc. Computer Vision and Pattern Recognition (CVPR), IEEE}
  (2019).

\bibitem[TZN19]{thies2019neural}
\textsc{Thies J., Zollh{\"o}fer M., Nie{\ss}ner M.}:
\newblock Deferred neural rendering: Image synthesis using neural textures.
\newblock \emph{ACM Transactions on Graphics 2019 (TOG)} (2019).

\bibitem[TZT{\etalchar{*}}20]{thies2020imageguided}
\textsc{Thies J., Zollh{\"o}fer M., Theobalt C., Stamminger M., Nie{\ss}ner
  M.}:
\newblock Image-guided neural object rendering.
\newblock In \emph{International Conference on Learning Representations}
  (2020).

\bibitem[VRM{\etalchar{*}}18]{vogels2018denoising}
\textsc{Vogels T., Rousselle F., McWilliams B., R{\"o}thlin G., Harvill A.,
  Adler D., Meyer M., Nov{\'a}k J.}:
\newblock Denoising with kernel prediction and asymmetric loss functions.
\newblock \emph{ACM Transactions on Graphics (TOG) 37}, 4 (2018), 124.

\bibitem[WLZ{\etalchar{*}}18]{wang2018vid2vid}
\textsc{Wang T.-C., Liu M.-Y., Zhu J.-Y., Liu G., Tao A., Kautz J., Catanzaro
  B.}:
\newblock Video-to-video synthesis.
\newblock In \emph{Advances in Neural Information Processing Systems (NeurIPS)}
  (2018).

\bibitem[Won16]{raymarch}
\textsc{Wong J.}:
\newblock Ray marching and signed distance functions, 2016.
\newblock Accessed: 2019-01-12.

\bibitem[WXCT19]{DBLP:journals/corr/abs-1906-01689}
\textsc{Werhahn M., Xie Y., Chu M., Thuerey N.}:
\newblock A multi-pass {GAN} for fluid flow super-resolution.
\newblock \emph{CoRR abs/1906.01689} (2019).

\bibitem[WYY{\etalchar{*}}14]{wang2014automatic}
\textsc{Wang R., Yang X., Yuan Y., Chen W., Bala K., Bao H.}:
\newblock Automatic shader simplification using surface signal approximation.
\newblock \emph{ACM Transactions on Graphics (TOG) 33}, 6 (2014), 1--11.

\bibitem[WZZH18]{wu2018fast}
\textsc{Wu H., Zheng S., Zhang J., Huang K.}:
\newblock Fast end-to-end trainable guided filter.
\newblock In \emph{Proceedings of the IEEE Conference on Computer Vision and
  Pattern Recognition} (2018), pp.~1838--1847.

\bibitem[XFCT18]{xie2018tempoGAN}
\textsc{Xie Y., Franz E., Chu M., Thuerey N.}:
\newblock tempogan: A temporally coherent, volumetric gan for super-resolution
  fluid flow.
\newblock \emph{ACM Transactions on Graphics (TOG) 37}, 4 (2018), 95.

\bibitem[YB18]{DBLP:journals/cgf/YangB18}
\textsc{Yang Y., Barnes C.}:
\newblock Approximate program smoothing using mean-variance statistics, with
  application to procedural shader bandlimiting.
\newblock \emph{Comput. Graph. Forum 37}, 2 (2018), 443--454.

\bibitem[ZIE{\etalchar{*}}18]{zhang2018perceptual}
\textsc{Zhang R., Isola P., Efros A.~A., Shechtman E., Wang O.}:
\newblock The unreasonable effectiveness of deep features as a perceptual
  metric.
\newblock In \emph{CVPR} (2018).

\bibitem[ZJL{\etalchar{*}}15]{zwicker2015recent}
\textsc{Zwicker M., Jarosz W., Lehtinen J., Moon B., Ramamoorthi R., Rousselle
  F., Sen P., Soler C., Yoon S.-E.}:
\newblock Recent advances in adaptive sampling and reconstruction for monte
  carlo rendering.
\newblock In \emph{Computer Graphics Forum} (2015), vol.~34, Wiley Online
  Library, pp.~667--681.

\end{thebibliography}
